\documentclass{article}

\usepackage{microtype}

\usepackage{hyperref}
\usepackage{url}
\usepackage{graphicx}
\usepackage{booktabs}       
\usepackage{amsfonts}       
\usepackage{nicefrac}       
\usepackage{amsmath}
\usepackage{microtype}      
\usepackage{xcolor}         
\usepackage{xpatch}
\usepackage{booktabs}
\usepackage{xpatch}
\usepackage[table]{xcolor}
\usepackage{wrapfig}
\usepackage{amsfonts} 
\usepackage{subcaption}
\usepackage{algorithm}
\usepackage{bbding}
\usepackage{multirow}
\usepackage[table]{xcolor}
\usepackage{multicol}
\usepackage{subcaption}
\usepackage{soul}          
\usepackage{enumitem}
\usepackage{placeins} 
\usepackage[theorems,breakable]{tcolorbox}%
\definecolor{HL1}{HTML}{FFF2CC} 
\definecolor{HL2}{HTML}{E2F0D9} 
\definecolor{HL3}{HTML}{DDEBF7} 
\definecolor{HL4}{HTML}{FCE4D6} 
\definecolor{HL5}{HTML}{EAE1F7} 
\newcommand{\HL}[2]{{%
    \sethlcolor{#1}
    \hl{#2}%
}}

\usepackage[preprint]{icml2026}

\usepackage{amsmath}
\usepackage{amssymb}
\usepackage{mathtools}
\usepackage{amsthm}

\usepackage[capitalize,noabbrev]{cleveref}

\theoremstyle{plain}

\theoremstyle{definition}

\theoremstyle{remark}

\newcommand{\revise}[1]{\textcolor{black}{#1}}
\usepackage[textsize=tiny]{todonotes}

\icmltitlerunning{To Think or Not To Think, That is The Question for Large Reasoning Models in Theory of Mind Tasks}
\newtcbtheorem{lem}{}{%
    colback=orange!5,
    colframe=orange!65,
    fonttitle=\bfseries,
    title after break={Theorem  -- \raggedleft Continued},
    detach title,
    boxsep=1pt,   
    left=5pt,     
    right=5pt,    
    top=5pt,      
    bottom=5pt,   
}{lem}
\begin{document}

\twocolumn[
  \icmltitle{To Think or Not To Think, That is The Question for \\ Large Reasoning Models in Theory of Mind Tasks}



  \icmlsetsymbol{eq}{*}
  \icmlsetsymbol{co}{${\dagger}$}

  \begin{icmlauthorlist}
    \icmlauthor{Nanxu Gong}{asu,eq}
    \icmlauthor{Haotian Li}{msra}
    \icmlauthor{Sixun Dong}{asu}
    \icmlauthor{Jianxun Lian}{msra}
    \icmlauthor{Yanjie Fu}{asu}
    \icmlauthor{Xing Xie}{msra}
  \end{icmlauthorlist}

  \icmlaffiliation{asu}{Arizona State University, Tempe, Arizona, United States}
  \icmlaffiliation{msra}{Microsoft Research Asia, Beijing, China}

  \icmlcorrespondingauthor{Nanxu Gong}{nanxugong@outlook.com}
  \icmlcorrespondingauthor{Haotian Li}{haotian.li@microsoft.com}

  \icmlkeywords{Machine Learning, ICML}

  \vskip 0.3in
]




\printAffiliationsAndNotice{$^*$Work done during internship at Microsoft Research Asia.}  

\begin{abstract}
Theory of Mind (ToM) assesses whether models can infer hidden mental states such as beliefs, desires, and intentions, which is essential for natural social interaction. Although recent progress in Large Reasoning Models (LRMs) has boosted step-by-step inference in mathematics and coding, it is still underexplored whether this benefit transfers to socio-cognitive skills. We present a systematic study of nine advanced Large Language Models (LLMs), comparing reasoning models with non-reasoning models on three representative ToM benchmarks. The results show that reasoning models do not consistently outperform non-reasoning models and sometimes perform worse. A fine-grained analysis reveals three insights.
First, slow thinking collapses: accuracy significantly drops as responses grow longer, and larger reasoning budgets hurt performance. 
Second, moderate and adaptive reasoning benefits performance: constraining reasoning length mitigates failure, while distinct success patterns demonstrate the necessity of dynamic adaptation.
Third, option matching shortcut: when multiple choice options are removed, reasoning models improve markedly, indicating reliance on option matching rather than genuine deduction.
We also design two intervention approaches: Slow-to-Fast (S2F) adaptive reasoning and Think-to-Match (T2M) shortcut prevention to further verify and mitigate the problems. 
With all results, our study highlights the advancement of LRMs in formal reasoning (e.g., math, code) cannot be fully transferred to ToM, a typical task in social reasoning. We conclude that achieving robust ToM requires developing unique capabilities beyond existing reasoning methods.

\end{abstract}

\section{Introduction}
Theory of Mind (ToM) refers to the human capacity to infer the unobservable mental states of others, such as beliefs, desires, and emotions, forming the foundation of social cognition \citep{survey_assess_enhance,survey1,survey2}.  It enables individuals to interpret subtle cues, anticipate behavior, and maintain meaningful communication. 
Research in large-scale reasoning has recently gained significant momentum \citep{deepseek-r1,qwen3}. Breakthroughs in Large Reasoning Models (LRMs) have shown that enhancing step-by-step inference capabilities yields dramatic improvements in structured domains like mathematics, code generation, and scientific problem-solving. These advances suggest that explicit reasoning serves as a general catalyst, potentially elevating models beyond surface-level pattern recognition toward more systematic and reliable intelligence. Against this backdrop, a critical question arises: Can these powerful reasoning mechanisms, proven effective in analytical domains, be successfully transferred to enhance the socio-cognitive capabilities required for ToM?

Existing research has only provided a preliminary exploration of the role of reasoning in ToM. The HiToM study demonstrated that applying Chain-of-Thought (CoT) prompting yields insignificant performance increases and may even amplify the model's susceptibility to deceptive information during testing \citep{HiToM}. Subsequent research has focused on incorporating more structured, human-like reasoning processes. Strategies such as integrating perspective-taking and reflection into CoT templates have effectively enhanced performance on ToM benchmarks \citep{FaR,SimToM(PT),MP}. More recently, significant attention has shifted towards Reinforcement Learning (RL) methods like GRPO, introduced by DeepSeek-R1, prompting exploration into R1-style inherent reasoning capabilities \citep{deepseek-r1,one-shot,gspo}. However, the efficacy of this approach remains contested; the ToM-RL study found that small models trained with GRPO can suffer from reasoning collapse, performing worse than those trained with standard Supervised Fine-Tuning (SFT) \citep{ToM-RL}. The current focus on method-level enhancements sidesteps a more fundamental question: 
\textit{From a model level, do reasoning models consistently outperform non-reasoning models in ToM?} A systematic analysis contrasting the ToM performance of reasoning models against non-reasoning models is essential to understand the effectiveness of reasoning capability and point out future directions.

To fill this gap, we conduct a comprehensive study on the effectiveness of LRMs on ToM tasks. Specifically, we evaluate model performance across three representative benchmarks, i.e., HiToM, ToMATO, and ToMBench, covering a range of reasoning order, taxonomy, and scenarios.  We find that reasoning models (e.g., DeepSeek-R1, Qwen3-8B-Reasoning) generally fail to outperform their non-reasoning counterparts (e.g., DeepSeek-V3, Qwen3-8B). For example, Qwen3-8B-Reasoning achieves a score of 0.648, which is significantly lower than the 0.705 scored by its non-reasoning version on the latest ToMATO benchmark. This counterintuitive performance suggests that the inherent reasoning capability in reasoning models may not be effective for ToM tasks. To move beyond this observation, we  provide a deeper analysis to diagnose when these failures occur and the underlying reasons why.

Our analysis reveals three fundamental insights behind models' reasoning errors:
\textbf{(i) Slow thinking correlates with reasoning collapse.}  We find that errors are heavily concentrated in longer responses, meaning the longer a model thinks, the more likely it is to fail. Pushing models like GPT-o3 and GPT-o4-mini to expend more reasoning effort actually backfires, leading to decreased performance. This demonstrates that for complex ToM tasks, prolonged computation is a liability, not an asset.
\textbf{(ii) Moderate and adaptive reasoning benefits performance.} We find that applying CoT to non-reasoning models and limiting the reasoning length of reasoning models both lead to performance gains. Furthermore, our experiments reveal the complementary strengths of reasoning and non-reasoning models. Consequently, we identify moderate and adaptive reasoning as a promising direction for future ToM research.
\textbf{(iii) Reasoning takes option matching rather than step-by-step deduction.}  Our experiments on HiToM demonstrate that when we remove the multiple-choice options, the performance of reasoning models such as DeepSeek-R1 and Qwen3-8B-Reasoning  dramatically improves. The study of their thinking process confirms why: they are not reasoning to a solution from the ground up, but rather matching the most likely answer from the provided list. This suggests their success often relies on option matching rather than genuine, step-by-step deduction. 
We also design and experiment LRMs with two intervention methods: \textbf{S}low-to-\textbf{F}ast reasoning (\textbf{S2F}) and \textbf{T}hink-to-\textbf{M}atch (\textbf{T2M}), which justifies that LRMs can perform better when the aforementioned issues are mitigated.

To summarize, the identified failure mechanisms of reasoning collapse and option matching shortcuts highlights the advancement of LRMs in formal reasoning (e.g., math, code) cannot effectively lead to increasing performance in ToM, a typical task in social reasoning. Our findings suggest that strategies beneficial in formal domains, such as prolonged deliberation, are often counterproductive in the ambiguous context of ToM. This implies that improving ToM is not about simply scaling existing analytical methods, but requires developing unique capabilities tailored for social reasoning. 
Our contributions can be summarized as:
\begin{itemize}[leftmargin=20pt]
    \item We provide a systematic comparison of reasoning and non-reasoning models on ToM tasks, revealing the counterintuitive finding that reasoning models  fail to establish advantages.

    \item We identify and provide empirical evidence for two core failure reasons in ToM reasoning: \textit{slow thinking collapse}, where prolonged deliberation becomes counterproductive, and \textit{option matching shortcut}, where models favor superficial pattern matching over genuine deduction. The findings are also verified with two intervention methods: Slow-to-Fast adaptive reasoning method and Think-to-Match shortcut prevention.

    \item Based on these findings, we discuss the fundamental divergence between ToM and formal reasoning and propose future directions.
\end{itemize}

\section{Related Work}
\paragraph{Theory of Mind Capability Evaluation.}
The benchmarks in ToM are mainly based on the Sally-Anne test, following a multi-choice format \citep{survey1,survey2}. ToMi extends ToM-bAbi via procedurally varied narratives and a systematic sweep over reality, memory, and first-/second-order belief queries, while HI-TOM pushes the envelope to fourth-order belief reasoning \citep{HiToM}. Moving beyond templated narratives, FANTOM introduces dialogue-mediated settings and explicitly targets “illusory ToM,” where responses appear correct yet violate underlying logical constraints \citep{FANToM}. In parallel, BigToM \citep{BigToM} and OpenToM \citep{OpenToM} broaden the mental-state taxonomy to include percepts, desires, and emotions. Domain-specific evaluations have also emerged: NegotiationToM integrates belief–desire–intention (BDI) reasoning within multi-round bargaining dialogues \citep{NegotiationToM}, and ToMBench pursues near-comprehensive ATOMS coverage with bilingual construction to mitigate pretraining contamination \citep{ToMBench}. Complementary efforts explore search- and generation-centric data creation, including A*-driven diversification in ExploreToM \citep{ExploreToM} and LLM–LLM self-play with information asymmetry in ToMATO \citep{ToMATO}. To comprehensively assess reasoning effectiveness, we evaluate on HiToM, ToMBench, and ToMATO, which together span higher-order belief depth, a broad mental-state taxonomy, and diverse evaluation scenarios.

\paragraph{Large Reasoning Model Evaluation.}
Benchmarks for LRMs span mathematics, formal logic, commonsense, code, and agentic interaction. Math suites range from contest-style and grade-school word-problem sets (MATH, GSM8K) to visually grounded mathematics (MathVista) and chart reasoning (ChartQA) \citep{math,cobbe2021training,lu2023mathvista,masry2022chartqa}. Logical-reasoning datasets cover deductive and abductive regimes (ProofWriter, FOLIO) and relational induction stress tests (CLUTRR) \citep{tafjord2021proofwriter,han2022folio,sinha2019clutrr}. Commonsense resources probe physical plausibility and broad knowledge (WinoGrande, MMLU) \citep{sakaguchi2020winogrande,hendrycks2021mmlu}. Code benchmarks emphasize exactness and executability, from function synthesis to real-world issue resolution (HumanEval, MBPP, SWE-bench) \citep{chen2021humaneval,austin2021mbpp,jimenez2023swebench}. Finally, web/embodied environments evaluate multi-step planning and tool use under interaction (Mind2Web, ALFWorld) \citep{liu2023mind2web,shridhar2021alfworld}. We systematically evaluate LRMs on ToM tasks, diagnose the failure modes underlying their reasoning errors, and outline directions for strengthening social reasoning.

\begin{table*}[t]
    \centering
    \caption{Overall results of all reasoning and non-reasoning models on three benchmarks}
    \resizebox{\textwidth}{!}{%
    \begin{tabular}{l ccc cc | cc cc}
        \toprule
        \multirow{2}{*}{\textbf{Dataset}}& \multicolumn{3}{c}{\textbf{GPT Family}} & \multicolumn{2}{c}{\textbf{DeepSeek Family}} & \multicolumn{2}{c}{\textbf{Qwen3-8B}} & \multicolumn{2}{c}{\textbf{Qwen3-32B}} \\
        \cmidrule(lr){2-4} \cmidrule(lr){5-6} \cmidrule(lr){7-8} \cmidrule(lr){9-10}
         & GPT-4o & GPT-o4-mini & GPT-o3 & DeepSeek-V3 & DeepSeek-R1 & Qwen3-8B & Qwen3-8B-Reasoning & Qwen3-32B & Qwen3-32B-Reasoning \\
        \midrule
        HiToM    & 0.607 & 0.547 & \textcolor{green!60!black}{\textbf{0.747}} & \textcolor{red!60!black}{\textbf{0.694}} & 0.549 & \textcolor{red!60!black}{\textbf{0.558}} & 0.481 & 0.586 & \textcolor{green!60!black}{\textbf{0.680}} \\
        ToMATO   & \textcolor{red!60!black}{\textbf{0.822}} & 0.792 & 0.817 & \textcolor{red!60!black}{\textbf{0.782}} & 0.749 & \textcolor{red!60!black}{\textbf{0.705}} & 0.648 & \textcolor{red!60!black}{\textbf{0.732}} & 0.714 \\
        ToMBench & 0.797 & 0.803 & \textcolor{green!60!black}{\textbf{0.818}} & 0.763 & \textcolor{green!60!black}{\textbf{0.801}} & 0.674 & \textcolor{green!60!black}{\textbf{0.729}} & 0.754 & \textcolor{green!60!black}{\textbf{0.775}} \\
        \bottomrule
    \end{tabular}
    }
    
    \label{tab:tom_series_cols}
    \vspace{-0.4cm}
\end{table*}
\section{Experiments}

\subsection{Experimental Setup}
\label{sec:setup}
To comprehensively study the effectiveness of reasoning, we select nine models including (i) \textbf{Reasoning Models:} GPT-o4-mini, GPT-o3, DeepSeek-R1, Qwen3-8B-Reasoning, and Qwen3-32B-Reasoning;
(ii) \textbf{Non-Reasoning Models:} GPT-4o, DeepSeek-V3, Qwen3-8B, and Qwen3-32B. 
Unless not adjustable, all models use the same settings: temperature 0, top-p 1, and a maximum output length of 2048 tokens.
\revise{Our selection spans both matched native reasoning and non-reasoning models \textit{within the same families} (e.g., GPT-o4-mini, GPT-o3, and GPT-4o), as well as \textit{the hybrid reasoning models} under reasoning and non-reasoning mode respectively (e.g., Qwen3-8B and Qwen3-8B-Reasoning). 
These models also cover both proprietary and open-source models for comprehensive investigation.
This design allows us to first study the impact of reasoning between reasoning and non-reasoning model pairs.
Then, the results can be synthesized to outline the common drawbacks of existing reasoning models and point out future directions for developing future ones.}

To systematically evaluate ToM, we select three specialized benchmarks, each designed to probe a distinct aspect of social reasoning. This ensures a multi-faceted assessment of model capabilities.
(i) \textbf{HiToM} \citep{HiToM} focuses on the depth of reasoning. It tests a model's ability to handle complex, multi-level recursive beliefs (from 0th to 4th-order) in narratives that include deceptive agents.
(ii) \textbf{ToMATO} \citep{ToMATO} assesses ToM in realistic, interactive contexts. It uses conversation-based scenarios between role-playing agents to test how well a model can infer mental states from dynamic, ongoing dialogues.
(iii) \textbf{ToMBench} \citep{ToMBench} provides broad taxonomic coverage. It systematically evaluates a wide range of distinct mental states, including beliefs, desires, emotions, and intentions, ensuring a comprehensive assessment of ToM abilities. All the experimental results are evaluated by the accuracy. Full results are available in Appendix \ref{app:result}.

\subsection{Experimental Results}

To intuitively compare reasoning and non-reasoning models, we present a side-by-side performance comparison of models from the same series.  This analysis investigates whether a focus on reasoning leads to universal improvements in ToM tasks. 
Table \ref{tab:tom_series_cols} reveals a counterintuitive pattern: reasoning models fail to consistently outperform their non-reasoning counterparts. This trend is consistent across model families. While reasoning models retain a clear advantage on ToMBench, this superiority is completely reversed on ToMATO, where non-reasoning models achieve higher scores in every pairing. Similarly, HiToM presents mixed results, with non-reasoning models winning the majority of comparisons. These findings highlight that reasoning models do not guarantee success and can incur performance costs in specific contexts, prompting further questions on why reasoning models fail. Detailed results are in Appendix \ref{app:result}.
\begin{lem}{\includegraphics[width=1em]{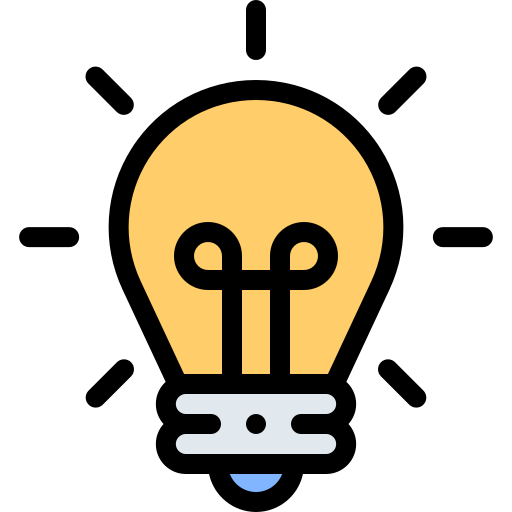}~\textbf{Takeaway 1}}{}%
\vspace{-0.2cm}
\includegraphics[width=1em]{img/light-bulb.png}~\textbf{Takeaway 1:} Reasoning models fail to demonstrate a definitive advantage over their non-reasoning counterparts across benchmarks.

\vspace{-0.1cm}
\end{lem}

\section{Analysis \& Intervention}
This section analyzes the overall experiment results in detail. It also introduces our additional experiments with designed interventions to verify and mitigate identified problems.

\subsection{Why Reasoning Models Fail to Outperform?}

\subsubsection{Response Length as a Signature of Failure}

\begin{figure}[htbp]
  \centering
  \begin{subfigure}[b]{0.235\textwidth}
    \centering
    \includegraphics[width=\textwidth]{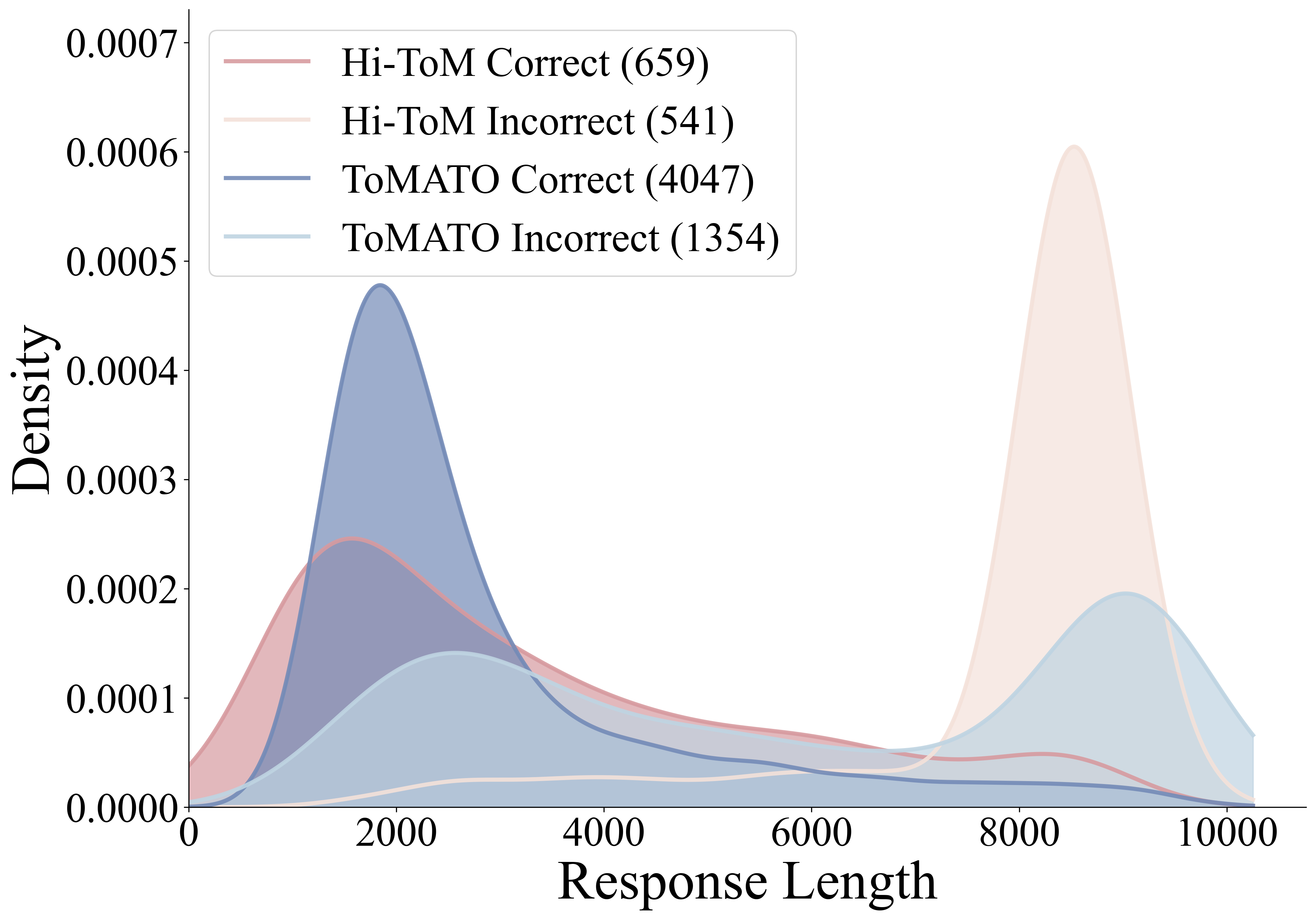}
    \subcaption{Response analysis}\label{fig:dis}
  \end{subfigure}
  \begin{subfigure}[b]{0.235\textwidth}
    \centering
    \includegraphics[width=\textwidth]{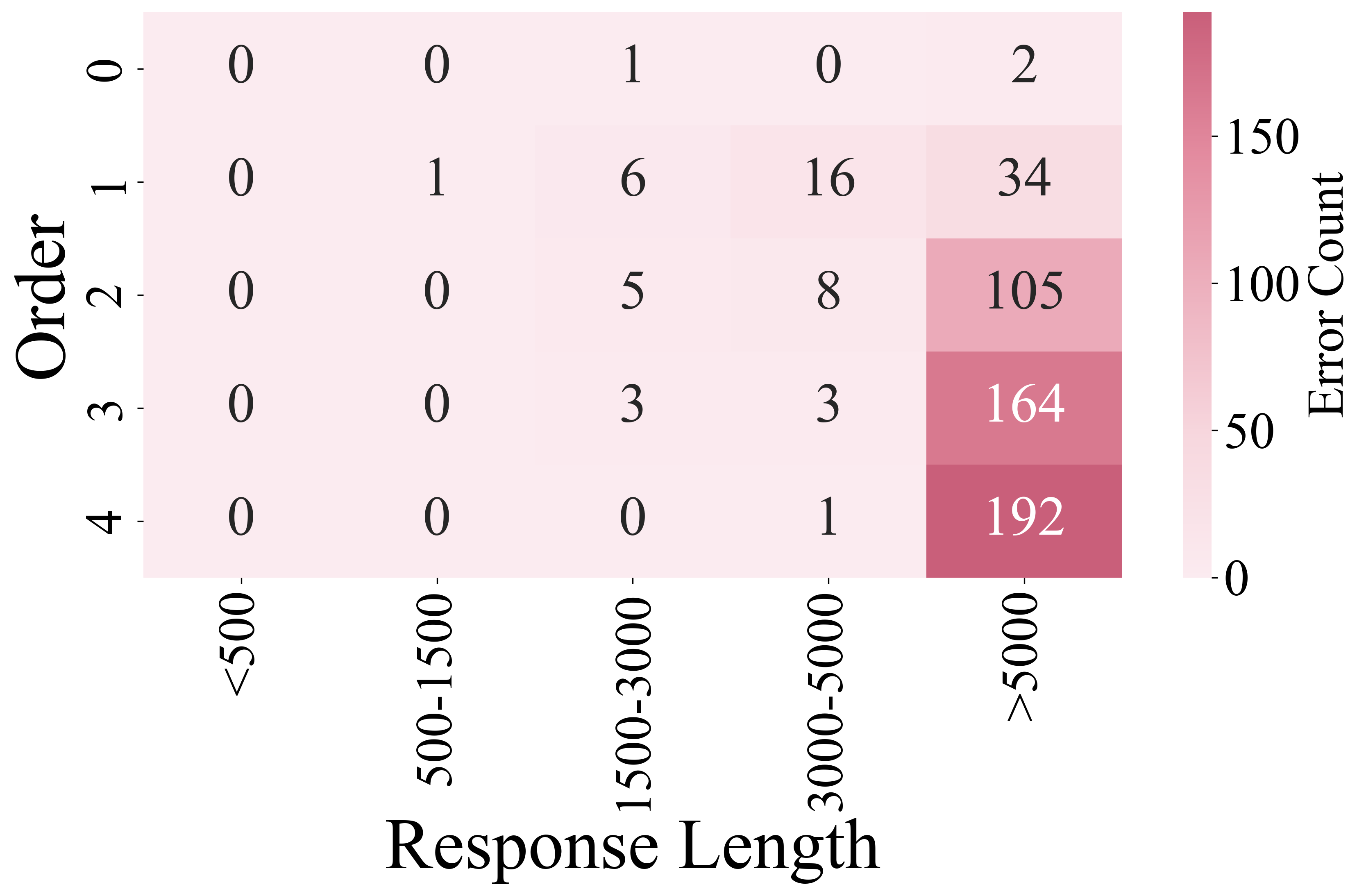}
    \subcaption{Orders and lengths in HiToM}\label{fig:heat}
  \end{subfigure}
  \caption{The distribution of the length and correctness of  reasoning model responses.  }
\vspace{-0.2cm}
\end{figure}

To understand how reasoning strategy changes between successful and failed attempts, we analyze the response length distribution of models. The results reveal a striking pattern, particularly on the HiToM benchmark. For example, as shown in Figure \ref{fig:dis}, the errors made by native reasoning model DeepSeek-R1 on this benchmark predominantly appear in a high-response-length region, forming a massive peak around 8,000 to 10,000 characters. However, this extreme pattern is mitigated on the ToMATO and ToMBench (Appendix \ref{app:rl_more}) benchmarks. While a distinction between the length of correct and incorrect responses for DeepSeek-R1 still exists on ToMATO, the separation is less pronounced, the error count is lower, and the distributions are more dispersed.
The reason for this difference becomes clear when we analyze the source of these failures. Figure \ref{fig:heat}, a heatmap of DeepSeek-R1's errors on HiToM, shows that the higher the task complexity, the more likely the model is to produce long and erroneous responses. While HiToM scales to the highly challenging Order 4, ToMATO peaks at the less demanding Order 2. 
Similarly, represented by Qwen3-8B-reasoning (Appendix \ref{app:rl_more} and \ref{app:heat_more}), the hybrid models also demonstrate the same pattern.
\revise{This suggests reasoning failure correlates with response length, especially on complex tasks.}
\label{sec:length}
\subsubsection{More Reasoning Does Not Lead to Better Performance}

The previous experiment suggests model reasoning response lengths correlate with failure, which may suggest extensive reasoning can backfire on the ToM performance.
To verify and explore the phenomenon, we conducted experiments on proprietary reasoning models and open-source reasoning models to explicitly control their reasoning for comparison.

\label{sec:effort}

\begin{figure}[h]
  \centering
  \begin{subfigure}[b]{0.23\textwidth}
    \centering
    \includegraphics[width=\textwidth]{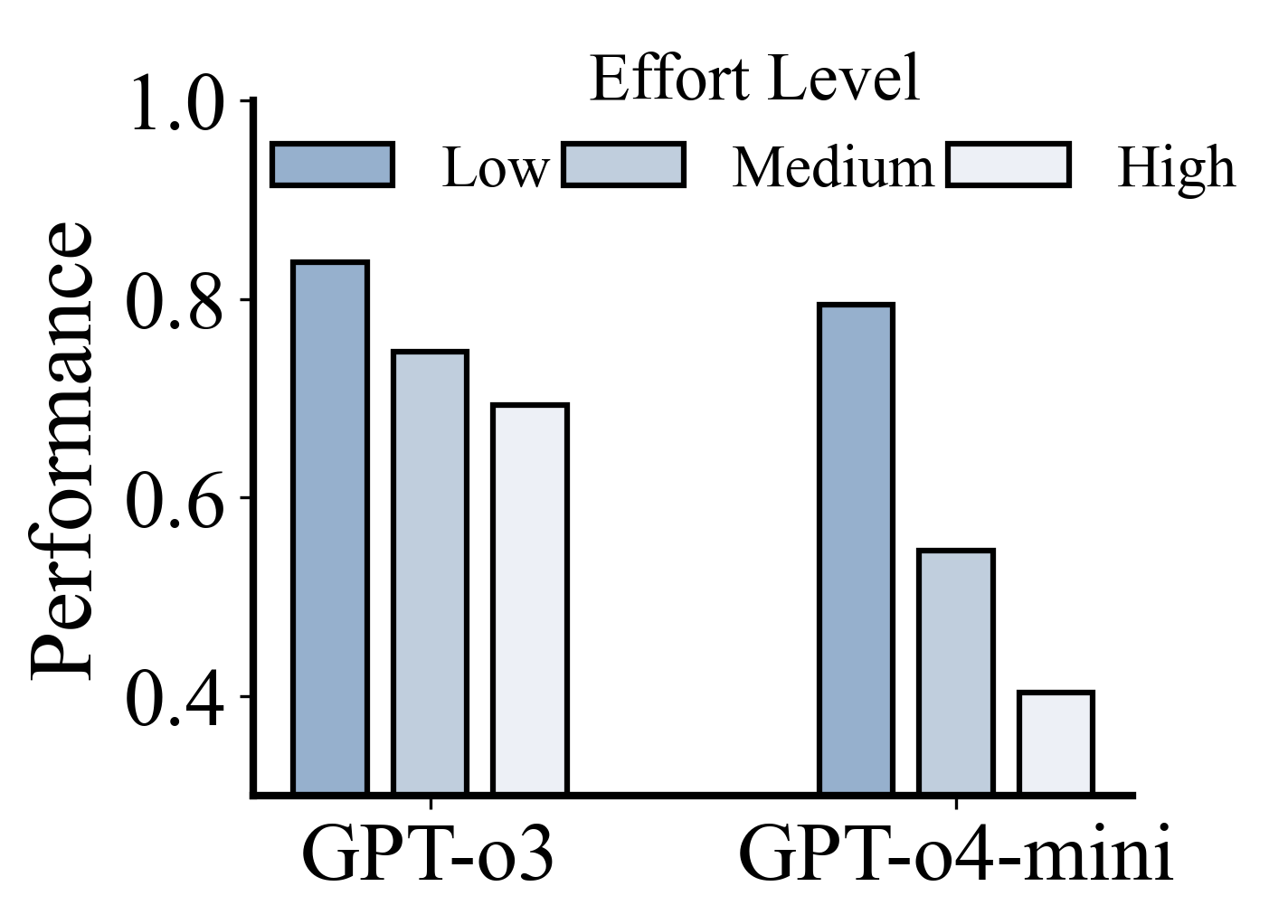}
    \subcaption{HiToM}\label{fig:bar}
  \end{subfigure}
  \begin{subfigure}[b]{0.23\textwidth}
    \centering
    \includegraphics[width=\textwidth]{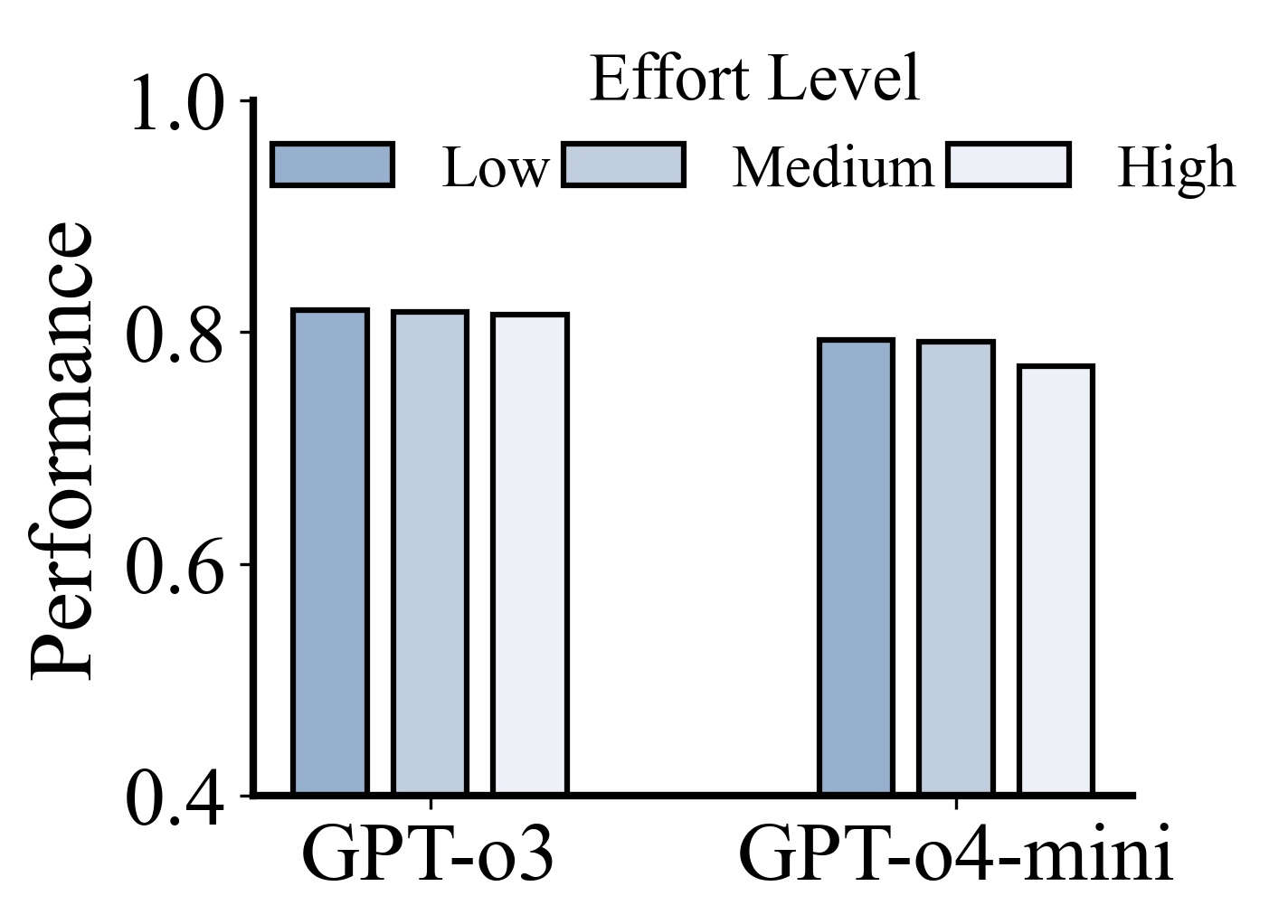}
    \subcaption{ToMATO}\label{fig:bar1}
  \end{subfigure}
  \label{fig:effort}
  \caption{Model performance with various reasoning efforts on benchmarks.}
  \vspace{-0.6cm}
\end{figure}

\textbf{Study of Proprietary Reasoning Models.} We control how much proprietary  models reason through a hyperparameter, \textit{reasoning effort}, of the GPT series used to control how much time and computational resources the model spends on internal Chain-of-Thought (CoT).
On the complex, higher-order HiToM benchmark, we observe a clear inverse relationship between effort and accuracy. As illustrated in Figure \ref{fig:bar}, the performance of the GPT-o3 model drops substantially from a high of 0.838 at the lowest effort level to 0.693 at the highest. This negative correlation is weakened on the less complex ToMATO benchmark (Figure \ref{fig:bar1}) and ToMBench (Appendix \ref{app:effort_tombench}). We found that varying the reasoning effort had a negligible impact on final accuracy. This divergence on two benchmarks demonstrates that the extensive reasoning failure is triggered by the high cognitive load of complex tasks.

\begin{figure}[htbp]
  \centering
  \captionsetup[subfigure]{font=tiny}
  \begin{subfigure}[t]{0.24\textwidth}
    \centering
    \includegraphics[width=\textwidth]{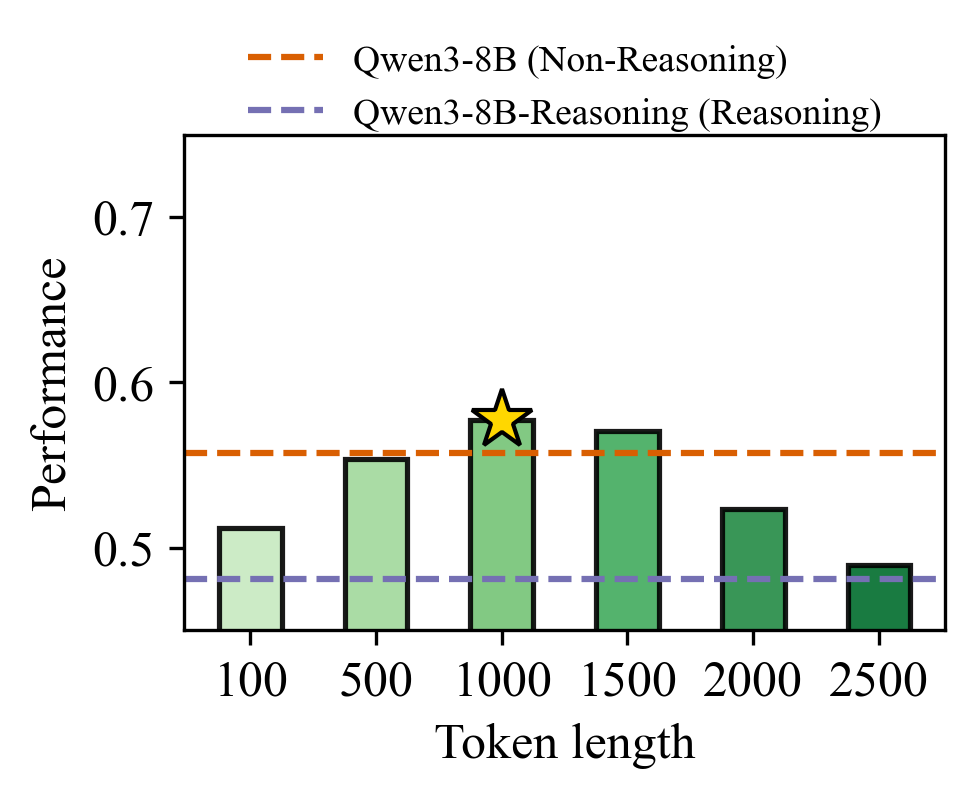}
    \caption{Qwen3-8B on HiToM}\label{fig:token}
  \end{subfigure}%
  \hfill
  \begin{subfigure}[t]{0.24\textwidth}
    \centering
    \includegraphics[width=\textwidth]{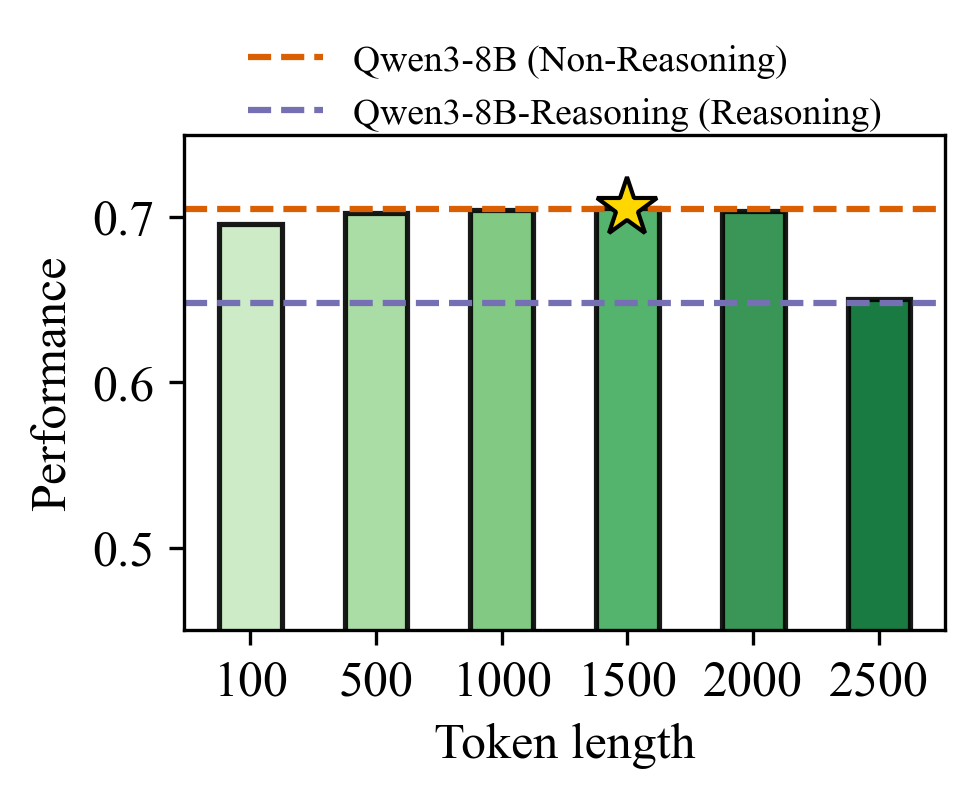}
    \caption{Qwen3-8B on ToMATO}\label{fig:token1}
  \end{subfigure}%
  \hfill
    \begin{subfigure}[t]{0.24\textwidth}
    \centering
    \includegraphics[width=\textwidth]{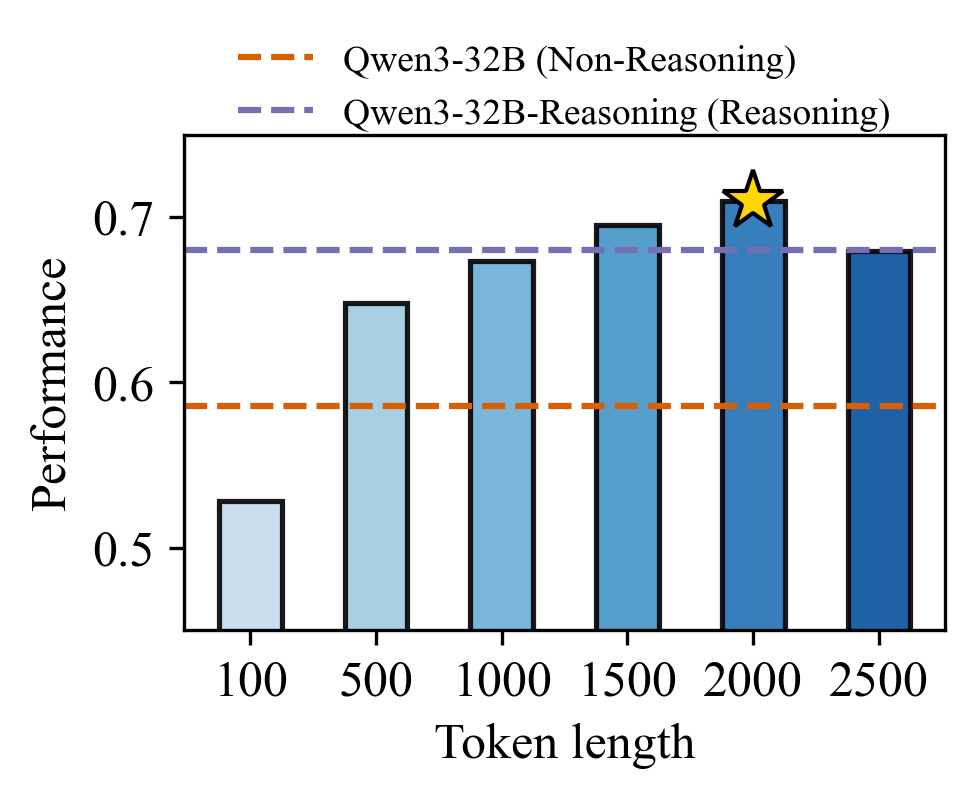}
    \caption{Qwen3-32B on HiToM}\label{fig:token2}
  \end{subfigure}%
  \hfill
  \begin{subfigure}[t]{0.24\textwidth}
    \centering
    \includegraphics[width=\textwidth]{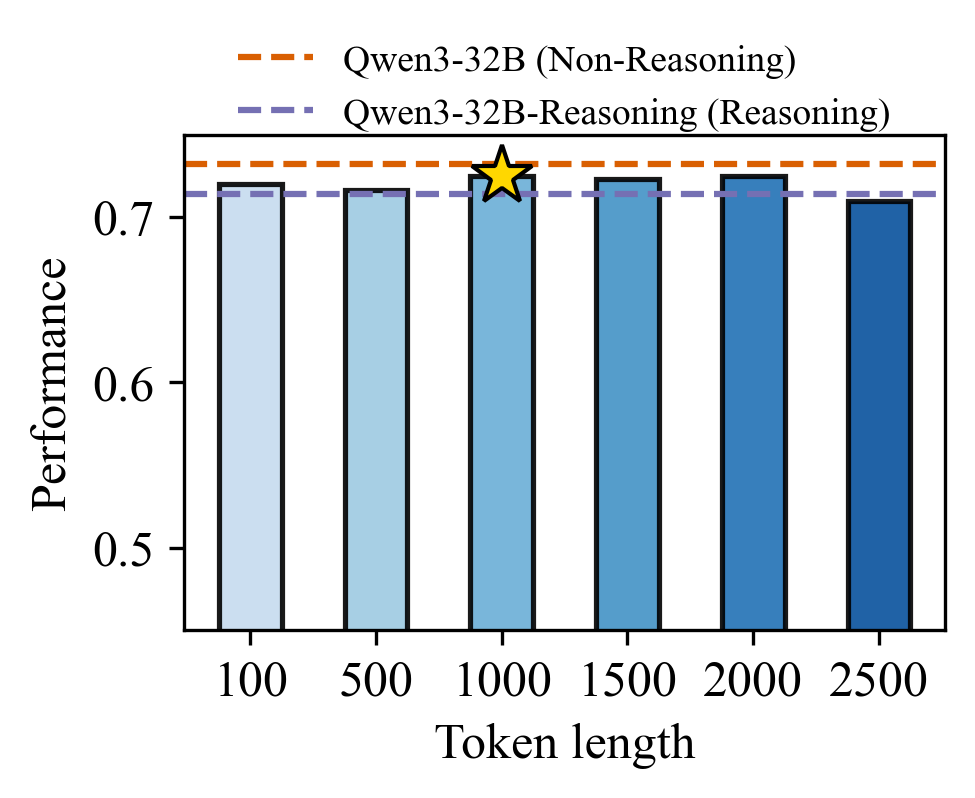}
    \caption{Qwen3-32B on ToMATO}\label{fig:token3}
  \end{subfigure}%
  \caption{Performance comparison under different token length limitations. Dash lines show original model performance without token limitation. Stars show token limits with best performance.}
  \label{fig:token_all}
    \vspace{-0.4cm}
\end{figure}

\begin{figure*}[ht]
    \centering
      \includegraphics[width=0.8\textwidth]{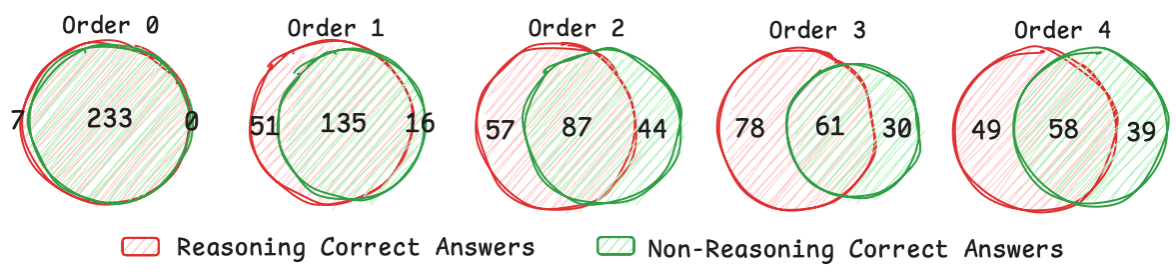}%
    \caption{Overlap between reasoning and non-reasoning models' correct answers. This example is from Qwen3-32B variants on HiToM.}
    \label{fig:overlap}
    \vspace{-0.4cm}
\end{figure*}

\textbf{Study of Open-source Reasoning Models.} On the open-source model side, we conduct an experiment by strictly limiting the reasoning tokens to further investigate the correlation between token length and performance.
Specifically, we set various thresholds for the maximum thinking length. When the length surpasses the threshold, we terminate the thinking process and append $</think>$ to the internal thought to force the LLM to derive an answer. 
We report the performance in Figure \ref{fig:token_all}. 
Interestingly, we find that imposing a token limit acts as a performance catalyst on complex HiToM tasks. For example, by restricting Qwen3-8B-Reasoning to think up to 1,500 tokens, we achieved a score of 0.706, surpassing both the non-reasoning and reasoning modes. Even on simpler benchmarks like ToMATO, managing token length yields results that are consistently superior to the model's native reasoning flow. For example, Qwen3-32B-Reasoning with 1,000 token limit achieves a score of 0.728 on ToMATO. In contrast, the unconstrained reasoning model achieves a lower score of 0.714, highlighting a clear performance advantage for the length-controlled approach.
Meanwhile, the results of an additional native reasoning model, R1-Distill-Qwen-7B, demonstrate the same pattern (Appendix \ref{app:token}). 
Furthermore, we identify that different models reach the best performance at different token length limits (e.g., on HiToM, Qwen-8B-Reasoning needs 1,000 tokens and Qwen-32B-Reasoning needs 2,000 tokens).
Even for the same model, the best token length limits also vary across benchmarks (e.g., Qwen-32B-Reasoning needs 2,000 tokens on HiToM and 1,000 tokens on ToMATO).
They suggest that we may not find an optimal token length limit number, implying the needs for adaptive reasoning length control to keep reasoning both effective and efficient.

\begin{lem}{\includegraphics[width=1em]{img/light-bulb.png}~\textbf{Takeaway 2}}{}%
\vspace{-0.2cm}
\includegraphics[width=1em]{img/light-bulb.png}~\textbf{Takeaway 2:} On complex ToM tasks, heavy reasoning expenditure, suggested by more thinking tokens or controlled by reasoning effort parameter, is associated with performance degradation rather than improvement.

\vspace{-0.1cm}
\end{lem}

\subsubsection{Moderate and Adaptive Reasoning Is Required for ToM Tasks}
\label{sec:complementary}
The experiments above justify that reasoning do not always bring benefits, especially when ToM tasks become more challenging.
We would like to further compare the performance of reasoning and non-reasoning models on tasks of different difficulties.
It can answer whether we need to fully abandon extensive reasoning for challenging ToM tasks.

The comparison of hybrid reasoning models provide a good testbed for examining the problem, as their reasoning and non-reasoning variants share same weights, implying a similar fundamental capability.
As an example, we show the comparison between Qwen3-32B and Qwen3-32B-Reasoning on HiToM in Figure \ref{fig:overlap}.
In low-complexity scenarios like Order 0 and Order 1, the models are largely in agreement, correctly answering a shared set of 233 and 135 samples, respectively. 
However, a significant divergence emerges at Order 2, where their complementary strengths become apparent with 87 overlapping correct answers but a combined 101 distinct correct answers where only one model succeeded. 
This trend culminates at the highest complexity, Order 4, where the Reasoning model uniquely solves 49 samples and the Non-Reasoning model uniquely solves 39, while they only agree on 58. 
The results indicate that extensive reasoning and no reasoning are complementary in addressing challenging ToM tasks.

The same finding holds for native reasoning and non-reasoning model comparison,  such as GPT-4o and GPT-o3 (Appendix \ref{app:overlap}). This substantial non-overlapping success clearly demonstrates that each model possesses unique problem-solving abilities that are crucial for tackling complex ToM challenges. \revise{These complementary strengths imply that neither pure reasoning nor non-reasoning models alone can achieve optimal results. Instead, an adaptive strategy that dynamically selects the appropriate reasoning budget based on task complexity offers a more promising path to superior performance.}

\begin{figure}[h]
  \centering
  \begin{subfigure}[b]{0.24\textwidth}
    \centering
    \includegraphics[width=\textwidth]{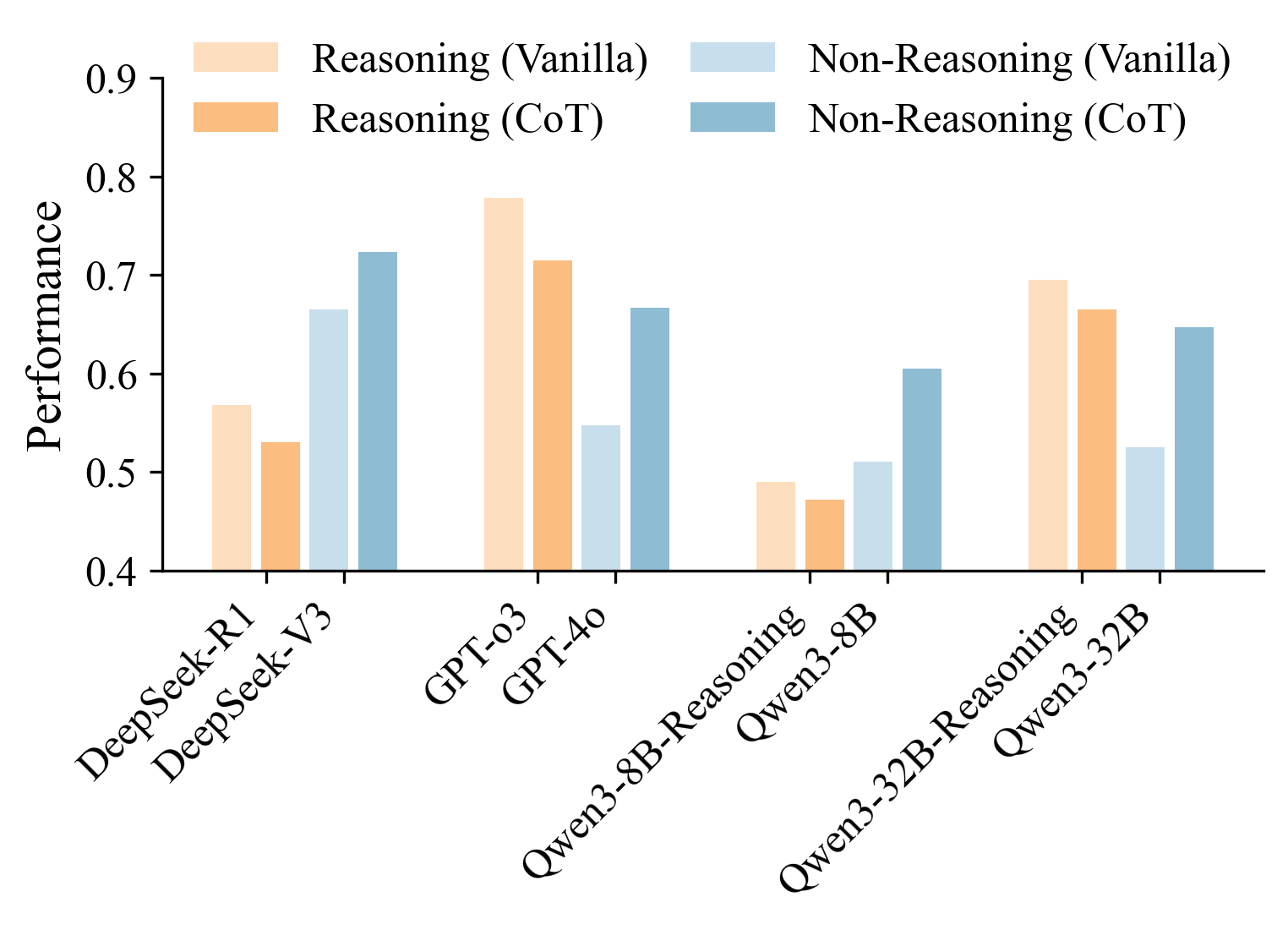}
    \subcaption{Model performance with CoT prompting}\label{fig:cot}
  \end{subfigure}
  \begin{subfigure}[b]{0.23\textwidth}
    \centering
    \includegraphics[width=\textwidth]{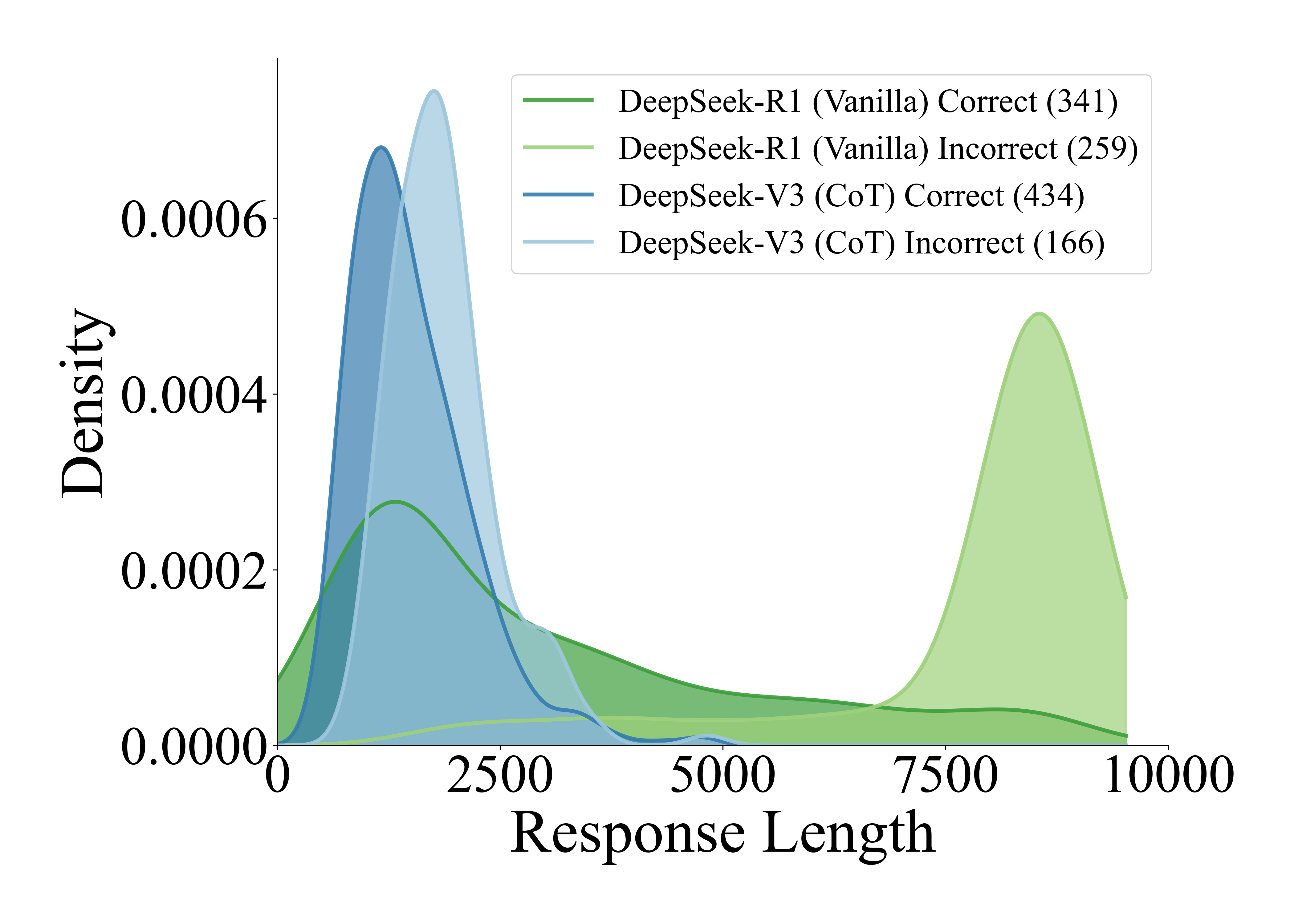}
    
    \subcaption{Response lengths and correctness }\label{fig:le}
  \end{subfigure}
  \caption{Model performance with CoT prompting. }
\vspace{-0.4cm}
\end{figure}

To validate the intuition of unleashing ToM capability with moderate and controlled reasoning, we conducted an additional experiment to prompt non-reasoning models with CoT prompts, simulating the reasoning process with mild efforts.
Figure \ref{fig:cot} shows that CoT prompting provides a significant boost to both native non-reasoning models and hybrid models under non-reasoning mode on the complex HiToM benchmark. For example, the performance of DeepSeek-V3 and Qwen3-32B increases substantially from 0.665 to 0.723 and from 0.525 to 0.647, respectively. Furthermore, our visualization of the response length distribution in Figure \ref{fig:le} also shows an interesting pattern. DeepSeek-V3 with CoT engages in moderate thinking, a process that improves its performance while successfully avoiding the pitfalls of prolonged deliberation and reasoning collapse. This balanced approach elevates the model's performance to nearly the level of the advanced GPT-o3. This suggests that while reasoning is inherently beneficial, the specific cognitive processes of current reasoning models may be flawed or misaligned with the unique demands of ToM tasks, which leads to the observed failure of reasoning. 

\begin{lem}{\includegraphics[width=1em]{img/light-bulb.png}~\textbf{Takeaway 3}}{}%
\vspace{-0.2cm}
\includegraphics[width=1em]{img/light-bulb.png}~\textbf{Takeaway 3:} Moderate and adaptive reasoning outperforms unconstrained slow thinking or no thinking, implying the needs for thinking strategy enhancement in ToM tasks.

\vspace{-0.1cm}
\end{lem}

\begin{figure*}[h]
  \centering
  \begin{subfigure}[b]{0.3\textwidth}
    \centering
    \includegraphics[width=\textwidth]{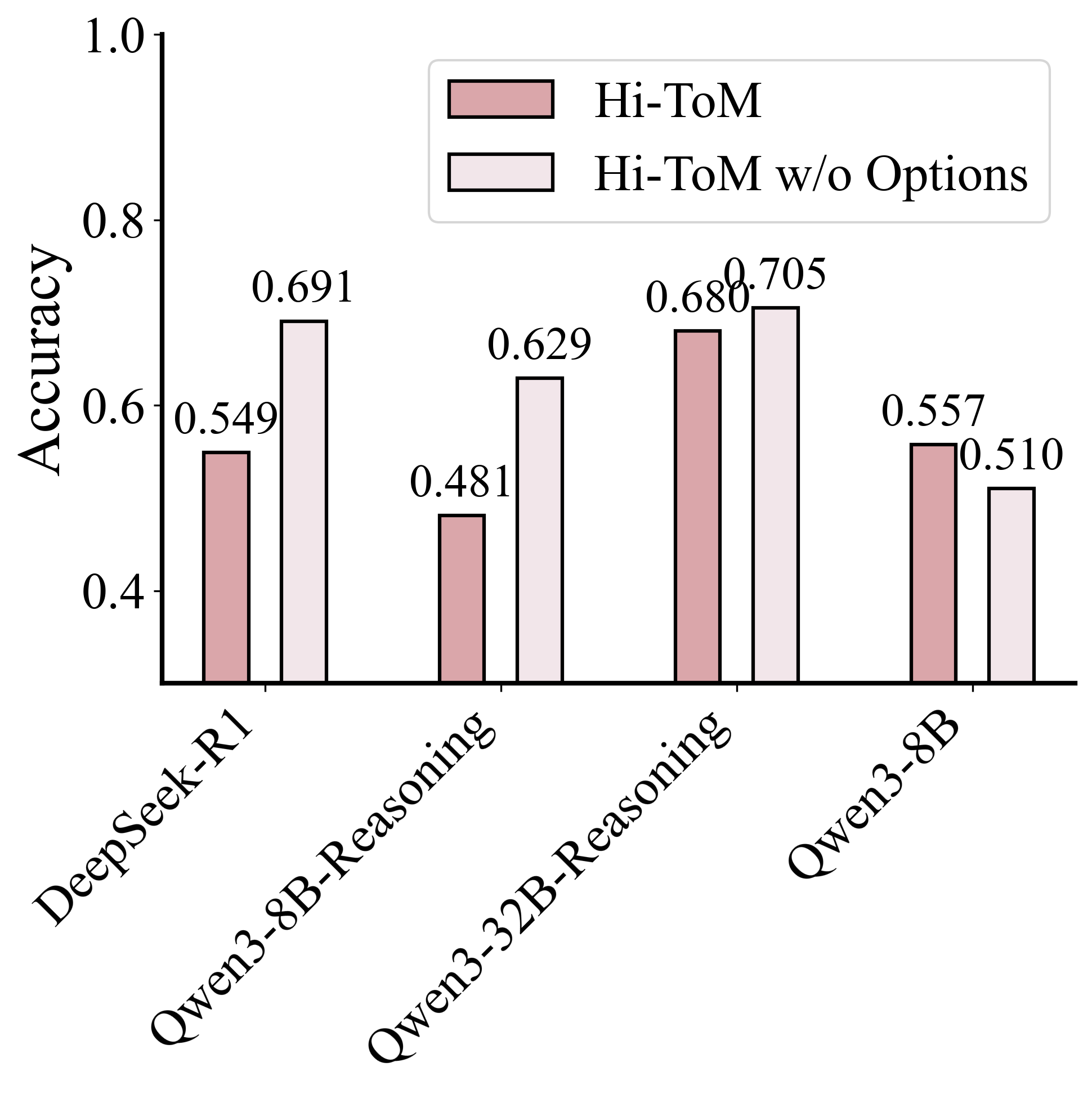}
    \subcaption{Accuracy comparison}\label{fig:op}
  \end{subfigure}
  \begin{subfigure}[b]{0.64\textwidth}
    \centering
    \includegraphics[width=\textwidth]{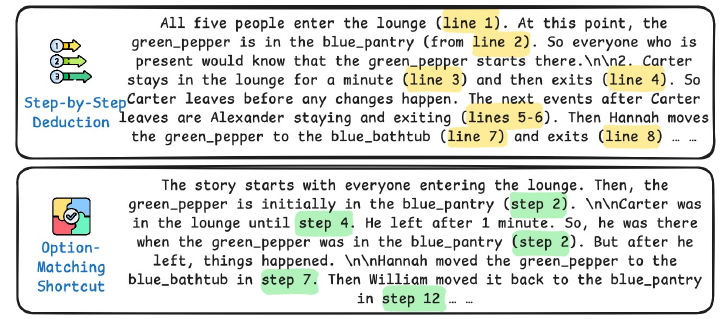}
    \subcaption{Behavior comparison with an example case}\label{fig:ex}
  \end{subfigure}

  \caption{The comparison of model performance when options are provided or not.}\label{fig:option}
  \vspace{-0.4cm}
\end{figure*}

\subsubsection{Option Hinders Successful Reasoning}
\label{sec:option}

When we probed into failure cases of reasoning models, we observed that models often relied heavily on the options, which led to a unnatural logic flow and final wrong answer.
As an example in Figure \ref{fig:ex} shows, when presented with options, reasoning models often abandon linear deduction, instead engaging in a chaotic search to find superficial justifications for potential choices. 
To better understand the influence of options on reasoning, we modified the HiToM benchmark by removing multiple-choice candidates, requiring models to extract answers directly. 
HiToM was selected specifically because its answers are explicitly extractive, ensuring unambiguous open-ended evaluation.
As shown in Figure \ref{fig:op}, removing options yields substantial improvements for both native reasoning models and hybrid models under reasoning mode. For example, DeepSeek-R1 surges from 0.549 to 0.691, and Qwen3-8B-Reasoning improves from 0.481 to 0.629. In contrast, the non-reasoning Qwen3-8B drops from 0.557 to 0.510. 
We also present the reasoning trace of the same model on the same question without options in Figure \ref{fig:ex}.
A structured deductive thinking process is demonstrated when the reasoning model is not aware of options.
These results suggest that explicit candidates short-circuit deduction, causing reasoning models to anchor on shallow overlaps. Meanwhile, non-reasoning baselines depend heavily on choice-level signals and lose support when those signals are removed.
\begin{lem}{\includegraphics[width=1em]{img/light-bulb.png}~\textbf{Takeaway 4}}{}%
\vspace{-0.2cm}
\includegraphics[width=1em]{img/light-bulb.png}~\textbf{Takeaway 4:} Explicit options lead to shortcuts in reasoning, causing models to prioritize superficial matching over deduction.

\vspace{-0.1cm}
\end{lem}

\subsection{\revise{Probing Reasoning Failure and Intervention Strategies in ToM}}

\revise{As we have identified a series of problems in ToM reasoning, we develop two intervention strategies to further verify and mitigate the problems in this section.} 
First, we leverage Slow-to-Fast reasoning to investigate the potential of an adaptive cognitive strategy. Building on it, we design Think-to-Match to deeply study the influence of the options on reasoning. Note that our experiments only include open-source models (e.g., Qwen3-8B, R1-Distill-Qwen-7B), due to the requirement of dynamic thinking control.

\subsubsection{Slow-to-Fast Reasoning}
\label{sec:S2F}
\begin{figure*}[h]
    \centering
    \resizebox{0.7
    \textwidth}{!}{%
      \includegraphics{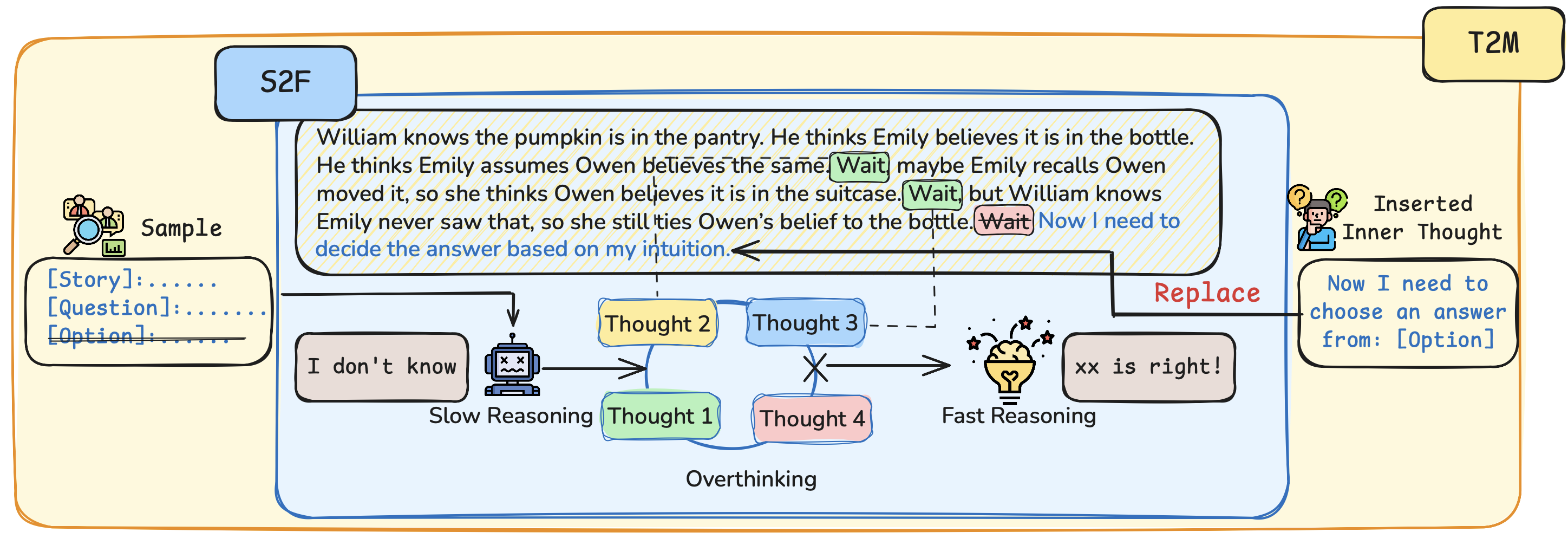}%
    }
    \caption{Overview of the Slow-to-Fast (S2F) and Think-to-Match (T2M) techniques. }
    \label{fig:method}
    \vspace{-0.4cm}
\end{figure*}

Our finding in Section~\ref{sec:effort} shows that we may not decide a unified token length limit for different reasoning models to achieve better ToM performance on various tasks.
As a result, we need an adaptive mechanism to decide when to stop thinking.
The existing works mark \textit{``wait''} as a proxy for extended deliberation. In reasoning models, such markers typically trigger additional verification or strategy switches, thereby lengthening the reasoning trajectory \citep{deepseek-r1}. 
To avoid prolonged slow thinking, inspired by previous research~\citep{alphaone}, S2F uses the frequency of \textit{``wait''} as a direct trigger for our dynamic intervention.
When its count exceeds a threshold, the overlong reasoning path tends to be unproductive in ToM tasks. Thus, we terminate the slow thinking process and compel the model to switch to fast thinking for an intuitive answer.

The results in Table \ref{tab:s2f} reveal that the effectiveness of S2F intervention is directly correlated with task complexity. The benefits are most pronounced on the complex, higher-order benchmark HiToM. For instance, the R1-Distill-Qwen-32B model’s performance surges from 0.571 to 0.701. However, on the 1st-order ToMBench, the intervention has a negligible impact across all models. These results offers a two-fold conclusion. First, the performance gains on HiToM demonstrate that the S2F strategy can successfully mitigate the problem of redundant reasoning.
It implies that most of the reasoning models cannot successfully determine the reasoning efforts for ToM questions and the problem is more severe in complex questions. 
They do not lack ToM capabilities, but their wrong reasoning strategy damages performance.
Furthermore, 
though our results verifies that moderate and adaptive thinking efforts can improve ToM performance, how to achieve the dynamic adjustment for thinking efforts requires more future research.
We notice that the model performance improvement led by S2F on ToMATO and ToMBench is less salient than the improvement on HiToM.
As discussed in Section \ref{sec:length}, the detrimental effects of slow thinking are less pronounced on simpler benchmarks, such as ToMATO. We suspect that the benefits gained from inhibiting slow thinking are counterbalanced by the disruption to the natural reasoning process, resulting in negligible overall improvement or even a slight decline in performance.
It suggests that the adaptive reasoning should be achieved by considering both deliberation efforts and question complexity, implying more future research on genuine ToM reasoning capability.

\begin{table*}[ht]
\centering
\caption{S2F performance across various reasoning models.}

\small
\resizebox{\textwidth}{!}{
\begin{tabular}{lcccccccccc}
\toprule
\multirow{2}{*}{\textbf{Benchmark}} & \multicolumn{2}{c}{\textbf{Qwen3-8B}} & \multicolumn{2}{c}{\textbf{Qwen3-32B}} & \multicolumn{2}{c}{\textbf{R1-Distill-Qwen-7B}} & \multicolumn{2}{c}{\textbf{R1-Distill-Qwen-32B}} & \multicolumn{2}{c}{\textbf{R1-Distill-Llama-8B}}\\ 
\cmidrule(lr){2-3} \cmidrule(lr){4-5} \cmidrule(lr){6-7} \cmidrule(lr){8-9} \cmidrule(lr){10-11}
& Vanilla & S2F & Vanilla & S2F & Vanilla & S2F & Vanilla & S2F & Vanilla & S2F \\
\midrule
HiToM & 0.481 & 0.557 \textcolor{green!60!black}{(+15.8\%)} &
0.680 & 0.682 \textcolor{green!60!black}{(+0.3\%)} &
0.353 & 0.397 \textcolor{green!60!black}{(+12.5\%)} &
0.571 & 0.701 \textcolor{green!60!black}{(+22.8\%)} &
0.396 & 0.451 \textcolor{green!60!black}{(+13.9\%)} \\
ToMATO & 0.648 & 0.700 \textcolor{green!60!black}{(+8.0\%)} &
0.714 & 0.724 \textcolor{green!60!black}{(+1.4\%)} &
0.490 & 0.505 \textcolor{green!60!black}{(+3.1\%)} &
0.706 & 0.708 \textcolor{green!60!black}{(+0.3\%)} &
0.586 & 0.578 \textcolor{red!70!black}{(-1.4\%)} \\
ToMBench & 0.729 & 0.731 \textcolor{green!60!black}{(+0.3\%)} &
0.775 & 0.777 \textcolor{green!60!black}{(+0.3\%)} &
0.559 & 0.560 \textcolor{green!60!black}{(+0.2\%)} &
0.773 & 0.769 \textcolor{red!70!black}{(-0.5\%)} &
0.655 & 0.626 \textcolor{red!70!black}{(-4.4\%)} \\
\bottomrule
\end{tabular}
}
\label{tab:s2f}
\end{table*}

\subsubsection{Think-to-Match}
\label{sec:t2m}
\begin{figure} 
  \centering
  \includegraphics[width=1\linewidth]{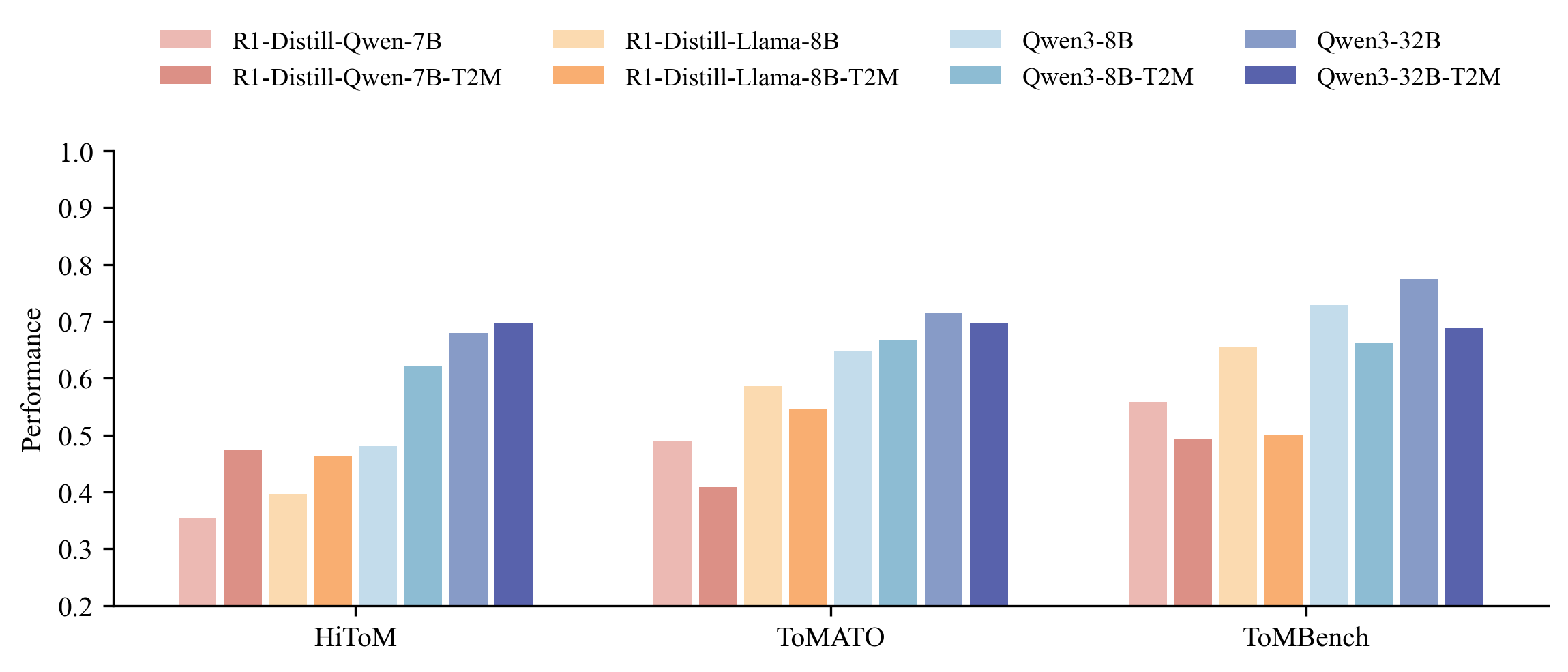}
  \caption{T2M performance.}
  \label{fig:t2m}
  \vspace{-0.6cm}
\end{figure}

Building on S2F, we introduce T2M to prevent option matching shortcuts. In the thinking phase, T2M removes answer options, compelling the model to perform first-principles reasoning constrained only by the S2F monitor. In the subsequent matching phase, options are reintroduced, prompting the model to align its generated deduction with the candidates. This approach ensures decisions are grounded in independent reasoning rather than superficial heuristics. As shown in Figure \ref{fig:t2m}, T2M significantly improves performance on HiToM, verifying that eliminating shortcuts unleashes the ToM capabilities of LRMs.

\begin{figure*}[h]
    \centering

    \resizebox{
    \textwidth}{!}{%
      \includegraphics{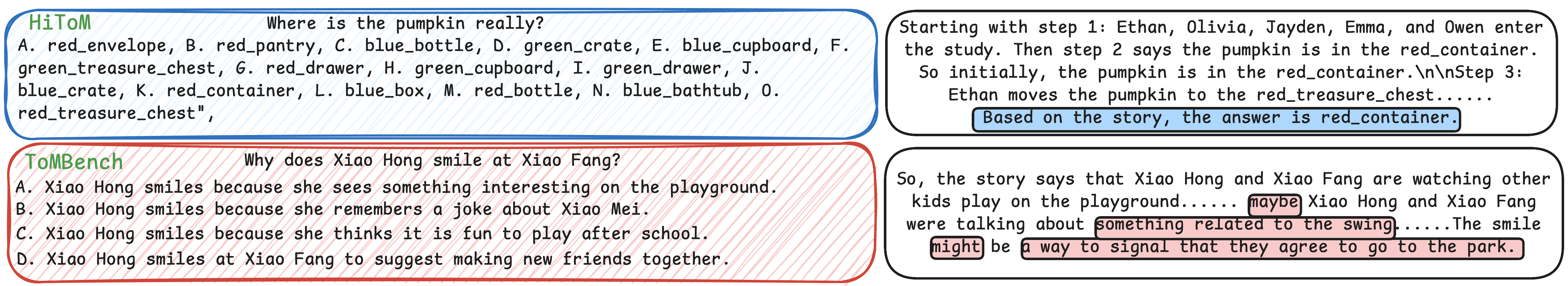}%
    }

    \caption{Cases to illustrate the different performance of T2M on HiToM and ToMBench.}
    \label{fig:case_op}
    \vspace{-0.4cm}
\end{figure*}

Furthermore, we learn an important lesson on ToM benchmark design from T2M's ineffectiveness on ToMATO and  ToMBench. Cases in Figure \ref{fig:case_op} reveal that the wrong results by T2M often follow the pattern: when the options do not match the initial reasoning direction, the answer can be wrong. 
This issue is related to the benchmark designs.
In ToMATO and ToMBench, questions are often more open-ended, such as inferring potential activities that a person might invite another to do.
These questions can have a large answer space, meaning that models can take various thinking directions to generate many more 
potentially correct answers than those provided as options.
As a result, without showing potential options in advance, T2M can lead to mismatch between the thinking directions and the provided options, further resulting in a wrong answer.
With the lens of T2M, a considerable challenge in ToM evaluation is revealed.
In a benchmark, questions with options can greatly restrict the search space for answers and thus lead to undesired option matching shortcuts. 
The performance growth led by these option matching behaviors is risky, as the ToM capability might be over-estimated.
On the other hand, if we only present the open-ended questions to models, the evaluation can be tricky, as the reasoning process may converge in many directions, leading to challenges of assessing them consistently.
Beyond the benchmark designs, it further validates that ToM reasoning heavily relies on options to guide thinking directions, implying that current LRMs may not fully obsess genuine deductive ToM reasoning capability.
It calls for efforts to develop dedicated training methods to enhance ToM reasoning in the future.

\subsection{Discussion}
Moving beyond the above experimental results, we derive key insights from the reasoning failure and provide inspiration for future directions in ToM reasoning method design.

\subsubsection{Insights from Reasoning Failure} 
\label{sec:insight}
We summarize our insights into 3 parts: \textit{1) Slow Thinking as a Liability:} Analysis (Sections \ref{sec:length} and \ref{sec:effort}) shows that extended deliberation in ToM tasks correlates with failure. Ambiguity triggers divergent search and counterproductive self-correction, which can be mitigated by constraining thinking tokens (Section \ref{sec:S2F}).
\textit{2) Bypassing Shortcut Biases:} Models often replace deduction with a "reverse-lookup" shortcut to justify provided options (Section  \ref{sec:option}). Enforcing a "think-then-match" strategy (Section  \ref{sec:t2m}) effectively counters this bias and improves distractor resistance.
\textit{3) Adaptive Reasoning Success:} Performance peaks when reasoning is moderate and adaptive (Sections \ref{sec:effort} and \ref{sec:S2F}). While constraints prevent cognitive collapse, dynamic transitions between reasoning and non-reasoning paradigms are essential for handling task complexity.

\subsubsection{From Formal to ToM Reasoning}

\textbf{Reasoning Effort Aids Formal Reasoning but Impairs ToM Reasoning.} ToM contrasts with formal reasoning by exhibiting a bi-phasic profile (Figure \ref{fig:dis}): correct answers are concise, while errors cluster in high-length regions. This indicates a ``reasoning collapse,'' where prolonged deliberation proves counterproductive. \revise{Traditionally, reasoning models show a distinct performance gap with non-reasoning models in formal reasoning. For example, DeepSeek-R1 performs clearly better than DeepSeek-V3 (e.g., math, code) \citep{deepseek-r1}. However, reasoning models fail to demonstrate dominated advantages on ToM benchmarks.} We attribute this to ToM's weakly verifiable nature. Unlike formal reasoning, where checkable intermediate states allow evidence to accumulate, extended ToM reasoning tends to amplify noise and induce perspective drift, often overwriting correct initial intuitions.

\textbf{Multiple Choice Helps Formal Reasoning, Hurts ToM Reasoning.} A second critical distinction lies in the impact of multiple-choice options. As we demonstrated in Figure \ref{fig:op}, for ToM reasoning tasks, the presence of an option set often degrades reasoning-model performance: models may adopt an option matching shortcut, becoming vulnerable to distractors. By contrast, in formal reasoning domains, a multiple-choice format can help by constraining the hypothesis space and providing clear targets for verification, often improving performance \citep{mc}. These divergent outcomes show that ToM and formal reasoning are qualitatively distinct problem classes requiring different capabilities and strategies.

\subsubsection{Towards ToM Reasoning Improvement}
\textbf{A System 1 \& System 2 Perspective.} Our findings suggest that progress on ToM reasoning can be fruitfully framed by dual-process theory, where non-reasoning models approximate an intuitive System 1 and reasoning models approximate a deliberative System 2 \citep{sys12,overthinking}. This view reconciles our results: Section \ref{sec:complementary} shows complementary strengths that neither system suffices universally. Section \ref{sec:effort} shows that a System 1-like model can be guided with simple CoT prompt to outperform System 2-like deliberate thinking. A key limitation is that current systems are typically run in a fixed mode per instance, rather than adapting effort to instance difficulty. Thus the goal is not merely building a stronger System 2, but integrating adaptive strategy selection. Future work can explore hybrids (e.g., a System 1 proposer with a System 2 verifier) and training that rewards choosing the minimal effective reasoning path. By shifting the target from thinking more to knowing how and when to think, we move toward more robust, human-like ToM reasoning.

\section{Conclusion}
This paper investigates whether LRMs can outperform non-reasoning models in ToM tasks, and find that their direct application is often ineffective and sometimes detrimental. Our systematic comparison reveals that reasoning models consistently fail to outperform their non-reasoning counterparts due to two primary failure reasons: reasoning collapse, where prolonged deliberation backfires, and a reliance on brittle option-matching shortcuts.
These findings highlight a fundamental divergence between the requirements of formal and social reasoning, showing that strategies successful in logic-based tasks become liabilities in ambiguous social contexts. This implies that improving ToM is not about simply scaling existing analytical methods but requires developing unique capabilities. Our intervention methods with Slow-to-Fast (S2F) reasoning and the Think-to-Match (T2M) confirm our findings and provide intervention solutions for the identified problems.
In the future, more research is required to develop LRMs with ToM capability, such as integrating System 1 and System 2 thinking through model training.

\section*{Impact Statement}
We acknowledge that ToM capabilities in AI models (e.g., LRMs) may carry potential risks, including deceptive manipulation, unwarranted anthropomorphism, and privacy encroachment. 
To understand and mitigate the risks, it is critical to gain more understanding of the related capabilities and advocate for appropriate usage based on the understanding.
Our work can serve as one step towards this direction.
For example, we observed the case of option-matching shortcut in LRMs' thinking process.
It warns future LRM usage.
Providing potential options or hints beforehand might lead to misleading answers for users.

\bibliography{Reference}
\bibliographystyle{icml2026}
\newpage
\appendix
\onecolumn
\appendix
\section{LLM Usage Statement}
We used LLMs (e.g., ChatGPT) only for grammar and wording edits.
\section{Reproducibility Statement}
We provide our hyperparameters for LLMs in Section \ref{sec:setup} and the pseudo-code of intervention methods in Section \ref{app:me}.
\section{Detailed Experimental Results}
\label{app:result}

\subsection{When Reasoning Models Fail to Outperform?}

\begin{table*}[htbp]
    \centering
    \caption{Performance of models on HiToM (reasoning orders) and ToMBench (taxonomy categories).}
    \resizebox{\textwidth}{!}{
    \begin{tabular}{lccccc|cccccc}
        \toprule
        \multirow{2}{*}{\textbf{Model}} & \multicolumn{5}{c}{\textbf{HiToM}} & \multicolumn{6}{c}{\textbf{ToMBench}}  \\ 
        \cmidrule(lr){2-6} \cmidrule(lr){7-12}
        & Order 0 & Order 1 & Order 2 & Order 3 & Order 4 & Belief & Desire & Emotion & Intention & Knowledge & Non \\
        \midrule
        \rowcolor{blue!10} \multicolumn{12}{c}{\textit{GPT Family}} \\
        GPT-o4-mini & \textcolor{green!60!black}{\textbf{1.000}} & 0.731 & 0.460 & 0.293 & 0.249 & 0.916 & 0.678 & 0.769 & 0.824 & \textcolor{green!60!black}{\textbf{0.648}} & 0.770 \\
        GPT-o3 & 0.996 & \textcolor{green!60!black}{\textbf{0.912}} & \textcolor{green!60!black}{\textbf{0.733}} & \textcolor{green!60!black}{\textbf{0.625}} & \textcolor{green!60!black}{\textbf{0.467}} & \textcolor{green!60!black}{\textbf{0.923}} & \textcolor{green!60!black}{\textbf{0.689}} & \textcolor{green!60!black}{\textbf{0.786}} & \textcolor{green!60!black}{\textbf{0.856}} & 0.617 & \textcolor{green!60!black}{\textbf{0.803}} \\
        GPT-4o & 0.979 & 0.692 & 0.571 & 0.408 & 0.383 & 0.909 & 0.661 & 0.762 & 0.847 & 0.624 & 0.782 \\
        \midrule
        \rowcolor{red!10} \multicolumn{12}{c}{\textit{DeepSeek Family}} \\
        DeepSeek-R1 & \textcolor{green!60!black}{\textbf{0.988}} & \textcolor{green!60!black}{\textbf{0.762}} & 0.508 & 0.292 & 0.196 & \textcolor{green!60!black}{\textbf{0.902}} & 0.661 & \textcolor{green!60!black}{\textbf{0.764}} & \textcolor{green!60!black}{\textbf{0.853}} & \textcolor{green!60!black}{\textbf{0.593}} & 0.794 \\
        DeepSeek-V3 & 0.979 & 0.650 & \textcolor{red!60!black}{\textbf{0.600}} & \textcolor{red!60!black}{\textbf{0.633}} & \textcolor{red!60!black}{\textbf{0.608}} & 0.812 & \textcolor{red!60!black}{\textbf{0.683}} & 0.738 & 0.838 & 0.479 & \textcolor{red!60!black}{\textbf{0.814}} \\
        \midrule \midrule
        \rowcolor{green!10} \multicolumn{12}{c}{\textit{Qwen3-8B}} \\
        Qwen3-8B-Reasoning & \textcolor{green!60!black}{\textbf{0.850}} & \textcolor{green!60!black}{\textbf{0.679}} & 0.421 & 0.246 & 0.208 & \textcolor{green!60!black}{\textbf{0.833}} & \textcolor{green!60!black}{\textbf{0.628}} & \textcolor{green!60!black}{\textbf{0.702}} & \textcolor{green!60!black}{\textbf{0.726}} & 0.479 & \textcolor{green!60!black}{\textbf{0.745}} \\
        Qwen3-8B & 0.846 & 0.667 & \textcolor{red!60!black}{\textbf{0.529}} & \textcolor{red!60!black}{\textbf{0.379}} & \textcolor{red!60!black}{\textbf{0.367}} & 0.656 & 0.611 & 0.674 & 0.715 & \textcolor{red!60!black}{\textbf{0.555}} & 0.737 \\
        \midrule 
        \rowcolor{yellow!10} \multicolumn{12}{c}{\textit{Qwen3-32B}} \\
        Qwen3-32B-Reasoning & \textcolor{green!60!black}{\textbf{1.000}} & \textcolor{green!60!black}{\textbf{0.775}} & \textcolor{green!60!black}{\textbf{0.600}} & \textcolor{green!60!black}{\textbf{0.579}} & \textcolor{green!60!black}{\textbf{0.446}} & \textcolor{green!60!black}{\textbf{0.872}} & 0.650 & \textcolor{green!60!black}{\textbf{0.743}} & 0.818 & \textcolor{green!60!black}{\textbf{0.534}} & \textcolor{green!60!black}{\textbf{0.782}} \\
        Qwen3-32B & 0.971 & 0.629 & 0.546 & 0.379 & 0.404 & 0.846 & \textcolor{red!60!black}{\textbf{0.689}} & 0.695 & \textcolor{red!60!black}{\textbf{0.824}} & 0.462 & 0.775 \\
        \bottomrule
    \end{tabular}
    }
    
    \label{tab:model_performance}
    \vspace{-0.2cm}
\end{table*}

\begin{table}[htbp]
\centering
\caption{ToMATO fine-grained accuracy by order and taxonomy}
\label{tab:tomato_finegrained}
\resizebox{\textwidth}{!}{
\begin{tabular}{lccccc|ccccc|ccccc}
\toprule
 \multirow{2}{*}{Model}& \multicolumn{5}{c}{1st-Order} & \multicolumn{5}{c}{2nd-Order} & \multicolumn{5}{c}{All} \\
 & belief & desire & emotion & intention & knowledge & belief & desire & emotion & intention & knowledge & belief & desire & emotion & intention & knowledge \\
\midrule
Qwen3-32B & 0.773 & 0.848 & 0.772 & 0.802 & 0.747 & 0.599 & 0.731 & 0.727 & 0.658 & 0.680 & 0.685 & 0.782 & 0.749 & 0.735 & 0.714 \\
Qwen3-32B-Reasoning & 0.773 & 0.859 & 0.788 & 0.751 & 0.696 & 0.586 & 0.724 & 0.713 & 0.617 & 0.663 & 0.679 & 0.783 & 0.750 & 0.689 & 0.680 \\
Qwen3-8B & 0.717 & 0.830 & 0.760 & 0.810 & 0.686 & 0.595 & 0.703 & 0.705 & 0.660 & 0.606 & 0.656 & 0.759 & 0.732 & 0.741 & 0.646 \\
Qwen3-8B-Reasoning & 0.719 & 0.839 & 0.735 & 0.750 & 0.644 & 0.493 & 0.570 & 0.677 & 0.518 & 0.572 & 0.605 & 0.689 & 0.706 & 0.642 & 0.608 \\
GPT-4o & 0.844 & 0.901 & 0.848 & 0.859 & 0.808 & 0.767 & 0.836 & 0.812 & 0.744 & 0.807 & 0.805 & 0.864 & 0.830 & 0.805 & 0.807 \\
GPT-o3 & 0.866 & 0.874 & 0.867 & 0.859 & 0.814 & 0.737 & 0.827 & 0.864 & 0.720 & 0.758 & 0.801 & 0.848 & 0.865 & 0.795 & 0.786 \\
GPT-o4-mini & 0.851 & 0.879 & 0.818 & 0.849 & 0.808 & 0.695 & 0.806 & 0.774 & 0.672 & 0.772 & 0.773 & 0.838 & 0.796 & 0.767 & 0.790 \\
DeepSeek-R1 & 0.789 & 0.874 & 0.786 & 0.827 & 0.720 & 0.670 & 0.726 & 0.758 & 0.672 & 0.693 & 0.729 & 0.791 & 0.772 & 0.755 & 0.706 \\
DeepSeek-V3 & 0.826 & 0.863 & 0.804 & 0.832 & 0.751 & 0.706 & 0.794 & 0.796 & 0.713 & 0.750 & 0.766 & 0.824 & 0.800 & 0.777 & 0.750 \\
\bottomrule
\end{tabular}}
\end{table}

Moving beyond overall performance comparison, this section provides a fine-grained analysis along two axes: the reasoning order and  taxonomy.  The benchmarks test different reasoning complexities, as ToMBench requires 1st-order reasoning, ToMATO up to 2nd-order, and HiToM up to 4th-order. Their taxonomic scope also expands from the belief-focused HiToM to the more comprehensive ToMBench, which covers 6 mental state dimensions. Our results confirm that model proficiency is not uniform: it degrades with higher-order reasoning and varies across different mental states.
To simplify our discussion, the remaining discussion will be mainly around ToMBench for comparing different ToM dimensions and HiToM for discussing orders (Table \ref{tab:model_performance}) .
The findings are also supported by our results on ToMATO (Table \ref{tab:tomato_finegrained}).

\subsubsection{Reasoning Loses Dominance in High-Order Inference.}

Table \ref{tab:model_performance} shows the detailed performance on different reasoning orders. In low-complexity scenarios (Orders 0-1), reasoning models exhibit a general performance advantage. For example, DeepSeek-R1 achieves an accuracy of 0.988 on 1st-order tasks. However, this trend becomes less consistent as reasoning complexity increases to Orders 2-4, where certain non-reasoning models begin to demonstrate comparable or superior performance. A stark example can be seen in that DeepSeek-V3 achieves a robust score of 0.608 on 4th-order tasks, while the performance of the reasoning-focused DeepSeek-R1 collapses to 0.196. Similarly, the reasoning process does not facilitate consistently superior performance, as exemplified by the comparison between Qwen3-8B and Qwen-8B-reasoning. A more in-depth analysis of these behaviors is in Section \ref{sec:length}.

\subsubsection{Reasoning's Benefits Vary Across Taxonomy }
We also study the influence of taxonomy on the performance of LLMs, as shown in Table \ref{tab:model_performance}.  A consistent advantage for reasoning models emerges in categories requiring the inference of structured, propositional attitudes.  For example, in Belief and Intention, reasoning models and models under reasoning mode like GPT-o3 and Qwen3-8B-reasoning consistently outperform the non-reasoning models, suggesting explicit thinking path is particularly effective for tracking cognitive states. However, this performance gap diminishes when assessing some mental states.  Most notably, in the Desire category, the non-reasoning Qwen3-32B is tied for the top score at 0.689.  These results reveal that the improvements from current reasoning capabilities are selective, rather than universal.  While they may enhance some aspects of ToM, they fail to provide a discernible advantage in challenging inferences, such as desire.

\subsection{Identifying Reasoning Error Types}
\begin{table}[t]
\centering
\scriptsize
\caption{General, cross-benchmark error taxonomy with simplified definitions and examples that highlight faulty model thinking. Each row uses a distinct highlight color.} 
\label{tab:error_type}
\begin{tabular}{p{0.23\textwidth} p{0.29\textwidth} p{0.40\textwidth}}
\toprule
\textbf{Error Type} & \textbf{Description} & \textbf{Example} \\
\midrule
\textbf{Evidence Grounding Error} &
Prediction is not grounded on the evidence or cites an entity that is not supported by the given context. &
\textit{Thinking:} ``Choose \HL{HL1}{\texttt{red\_basket}}; it \HL{HL1}{sounds plausible}.'' \newline
\textit{Fault:} \texttt{red\_basket} is not in the evidence set; no justification links it to the task options. \\
\midrule
\textbf{State Tracking Error} &
Timeline or observability is tracked incorrectly. &
\textit{Thinking:} ``The \HL{HL2}{last move} puts it in X, so \HL{HL2}{Owen believes X}.'' \newline
\textit{Fault:} Owen had already left before the last move; belief should freeze at his last observation. \\
\midrule
\textbf{Perspective Attribution Error} &
Answers the wrong subject or wrong mental construct. &
\textit{Thinking:} ``\HL{HL3}{I (A) saw it} in Y; therefore \HL{HL3}{B thinks Y}.'' \newline
\textit{Fault:} Egocentric projection; the question asks for A's belief about B's belief, not A's own knowledge. \\
\midrule
\textbf{Discourse Misinterpretation} &
Speech acts or rhetorical cues are treated as factual updates. &
\textit{Thinking:} ``He \HL{HL4}{\emph{claimed}} it is in X; so \HL{HL4}{it is in X and others now know X}.'' \newline
\textit{Fault:} Claims do not change world state or shared knowledge without corroboration. \\
\midrule
\textbf{Commonsense \& Causal Error} &
Over-applies a generic script or inserts unsupported causality not warranted by context. &
\textit{Thinking:} ``\HL{HL5}{This looks like a negotiation}; the \HL{HL5}{goal is to persuade}, so the counterpart is cautious.'' \newline
\textit{Fault:} Injects a persuasion script and causal story absent from the evidence; mislabels the speaker's stance. \\
\bottomrule
\end{tabular}

\end{table}

To understand why models fail, we move from quantitative scoring to a qualitative analysis of their reasoning processes. By manually inspecting the chain-of-thought outputs for incorrect predictions,  we identified five recurring and distinct categories of reasoning failures, beside the option-related one reported in our main paper (Section~\ref{sec:option}). These categories, which form a general, cross-benchmark error taxonomy, are defined and exemplified in Table~\ref{tab:error_type}.

\textbf{Evidence Grounding Errors.} This is one of the most fundamental types of failure, where the model's reasoning path deviates from the provided context. An Evidence Grounding Error occurs when a model bases its conclusion on information that is not present in the evidence or makes a decision by citing an entity that cannot be factually supported by the scenario. As the example in Table~\ref{tab:error_type} shows, this can manifest as the model choosing an option because it "sounds plausible" rather than because it is logically derived from the given facts. This error type highlights a critical weakness in a model's ability to strictly adhere to its input context, often resorting to unverified assumptions or hallucinated details.

\textbf{State Tracking Error.}
This error reveals a model's difficulty with temporal and observational logic. A State Tracking Error happens when the model incorrectly processes the timeline of events or fails to account for an agent's limited perspective. A common failure mode is updating an agent's belief based on an event that the agent did not witness. For instance, a model might incorrectly conclude that "Owen believes X" because the object was moved to location X, while failing to register that Owen had already left the room and could not have observed the move. This points to a deficient mechanism for maintaining and freezing the mental state of different agents at specific points in time.

\textbf{Perspective Attribution Error.}
This error is a classic failure of Theory of Mind, where the model fails to correctly simulate another agent's perspective and instead defaults to its own. A Perspective Attribution Error occurs when the model answers from the wrong point of view—often its own "all-seeing" one—or confuses whose mental state is being queried. This frequently manifests as egocentric projection, where the model imputes its own knowledge onto an agent (e.g., "I, the model, know it is in Y, therefore Agent B must think it is in Y"). This shows a breakdown in handling nested beliefs and maintaining the crucial distinction between objective reality and an agent's subjective perception.

\textbf{Discourse Misinterpretation.}
This category of error highlights a model's lack of pragmatic understanding in social communication. A Discourse Misinterpretation occurs when the model treats non-factual speech acts—such as claims, questions, jokes, or rhetorical statements—as literal updates to the world state or shared knowledge. For example, a model might incorrectly assume that because a character claimed an object was in a certain location, the object is now factually in that location and all other characters are aware of this. This reveals a naive, literal interpretation of language that misses the social nuances and reliability judgments inherent in human discourse.

\textbf{Commonsense \& Causal Error.}
This error type involves the misapplication of world knowledge or the fabrication of unsupported causal links. A Commonsense \& Causal Error happens when a model imposes a generic script or schema onto a situation where it does not fit, or when it invents a cause-and-effect relationship that is not warranted by the evidence. For example, a model might incorrectly classify a simple information exchange as a "negotiation," thereby misinterpreting the agents' stances and intentions. This shows that while models possess vast commonsense knowledge, they struggle to apply it appropriately and can over-generalize, leading to a distorted understanding of the specific context.

\subsection{Response Length Distribution}
\label{app:rl_more}
\begin{figure}[h]
  \centering
  \begin{subfigure}[b]{0.4\textwidth}
    \centering
    \includegraphics[width=\textwidth]{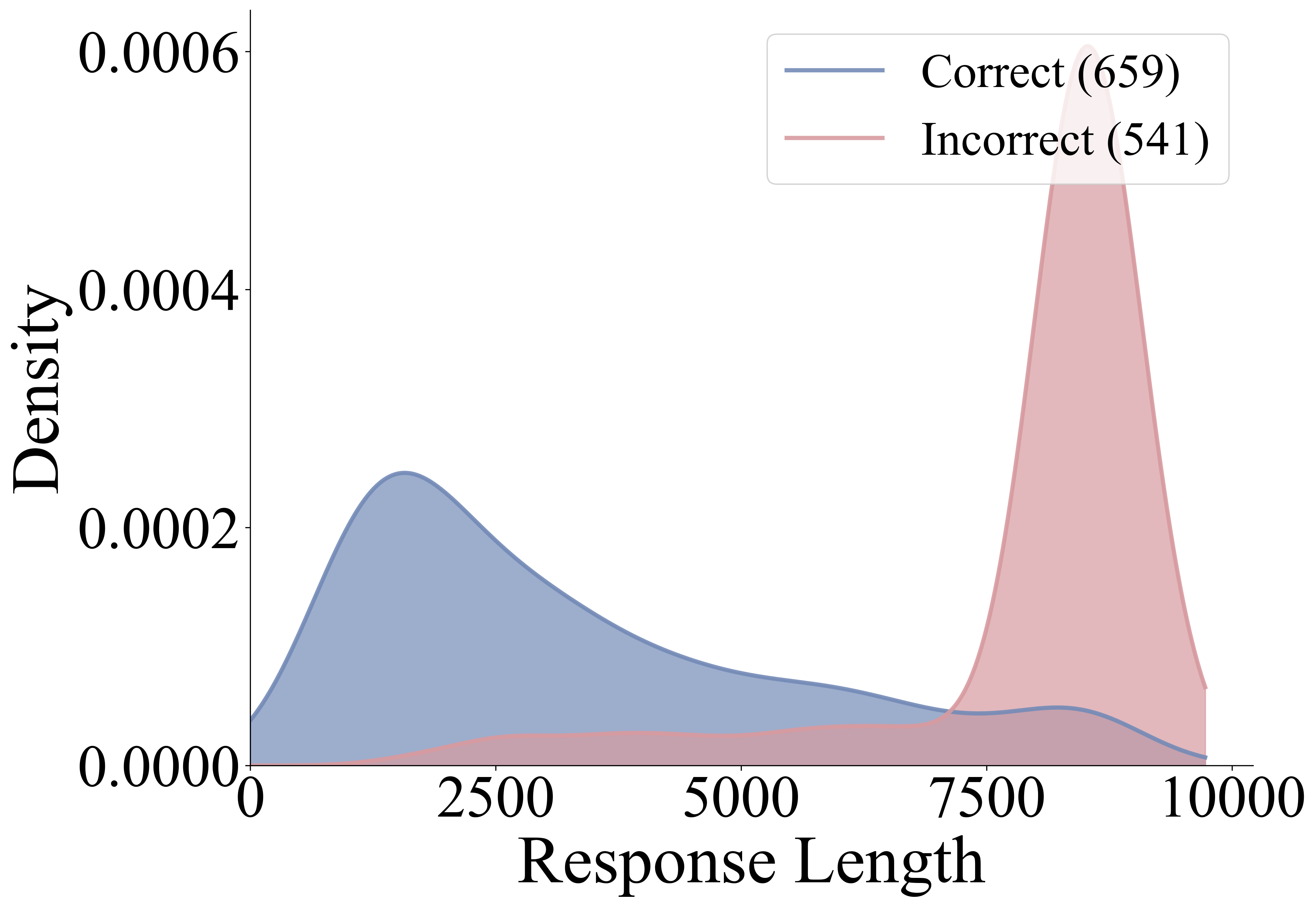}
    \subcaption{DeepSeek-R1}
  \end{subfigure}
  \begin{subfigure}[b]{0.4\textwidth}
    \centering
    \includegraphics[width=\textwidth]{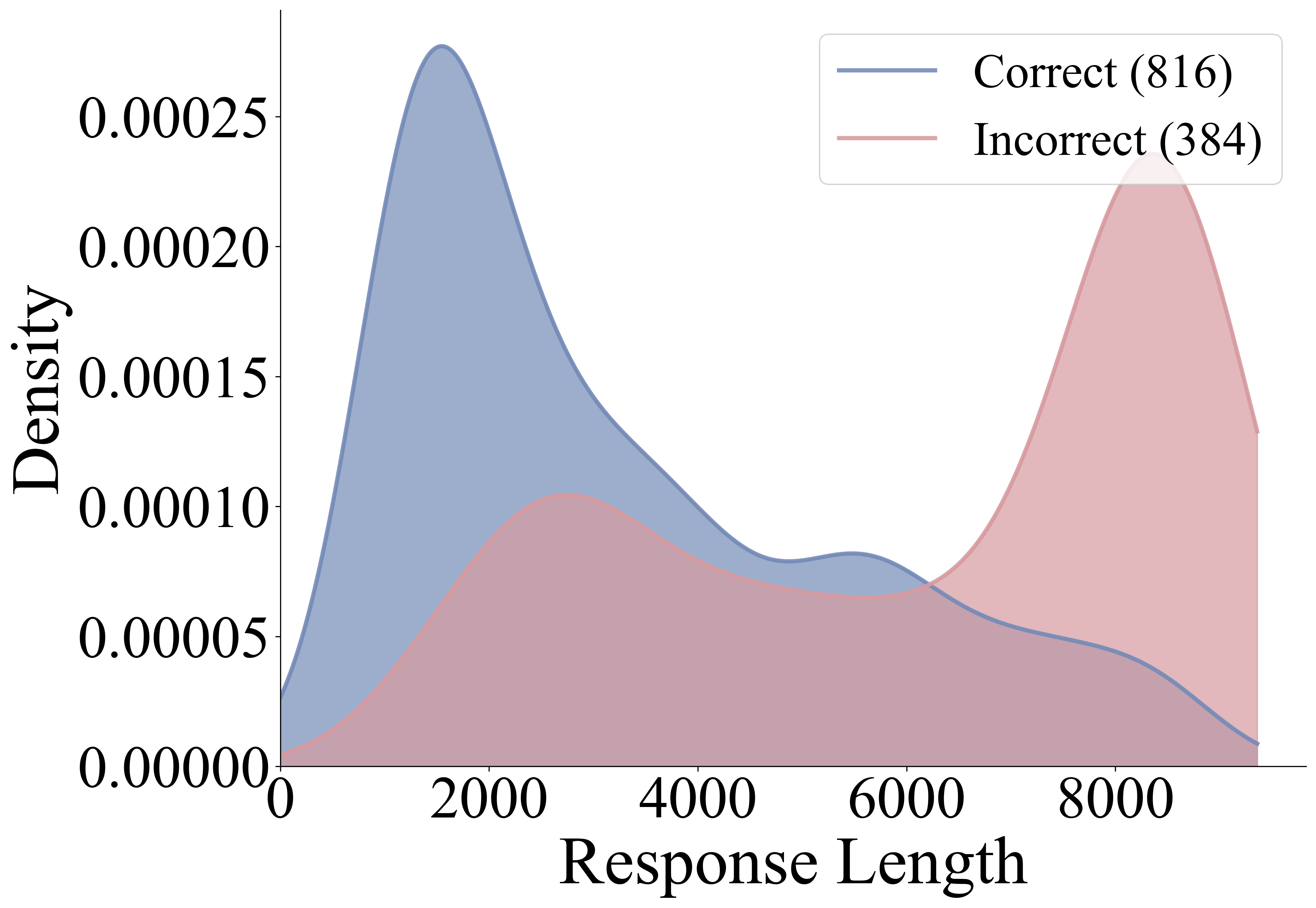}
    \subcaption{Qwen3-32B-Reasoning}
  \end{subfigure}
  \begin{subfigure}[b]{0.4\textwidth}
    \centering
    \includegraphics[width=\textwidth]{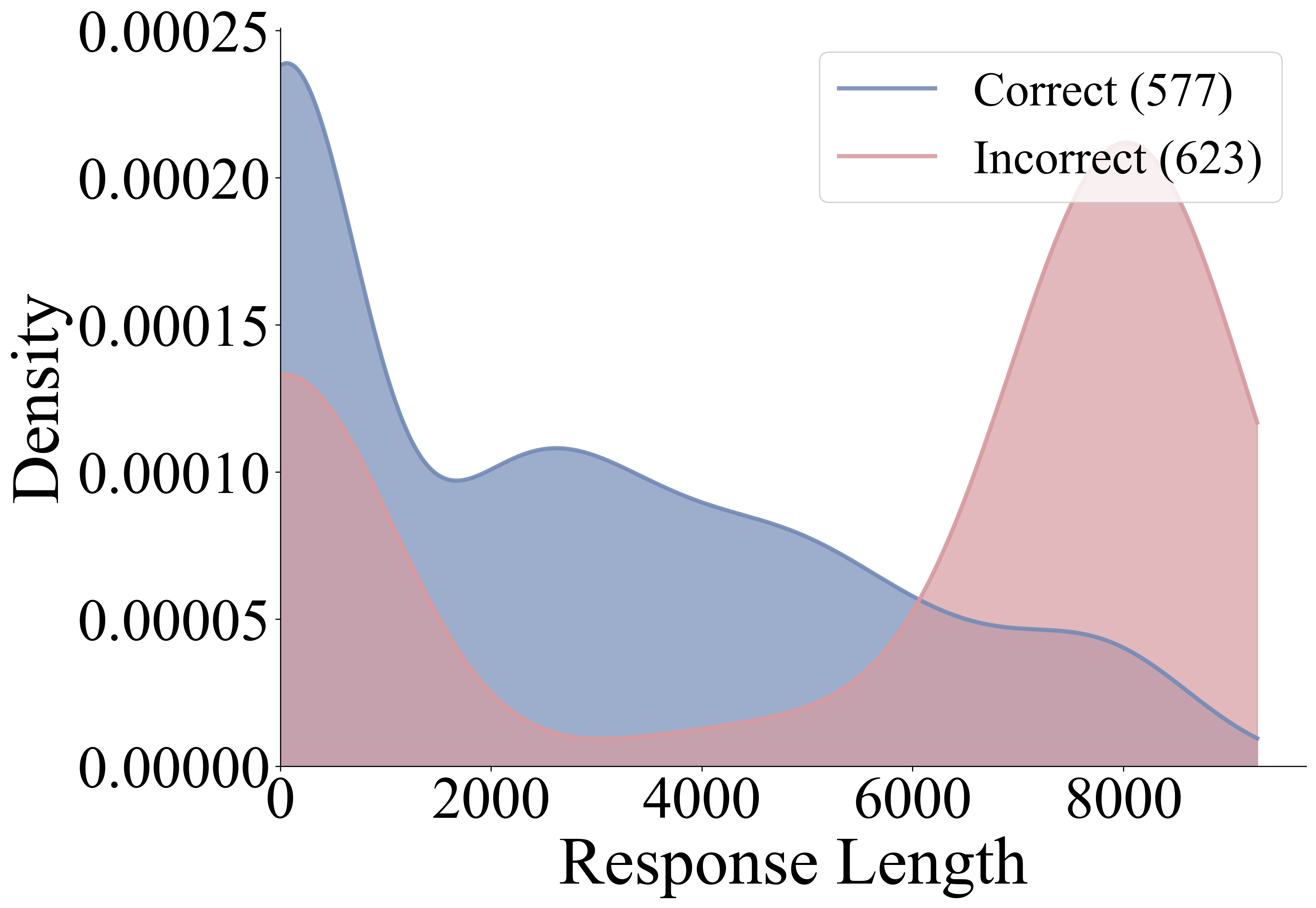}
    \subcaption{Qwen3-8B-Reasoning}
  \end{subfigure}
  \caption{Response Length Distribution on HiToM.}
\label{app:rl_hitom}
\end{figure}

\begin{figure}[h]
  \centering
  \begin{subfigure}[b]{0.4\textwidth}
    \centering
    \includegraphics[width=\textwidth]{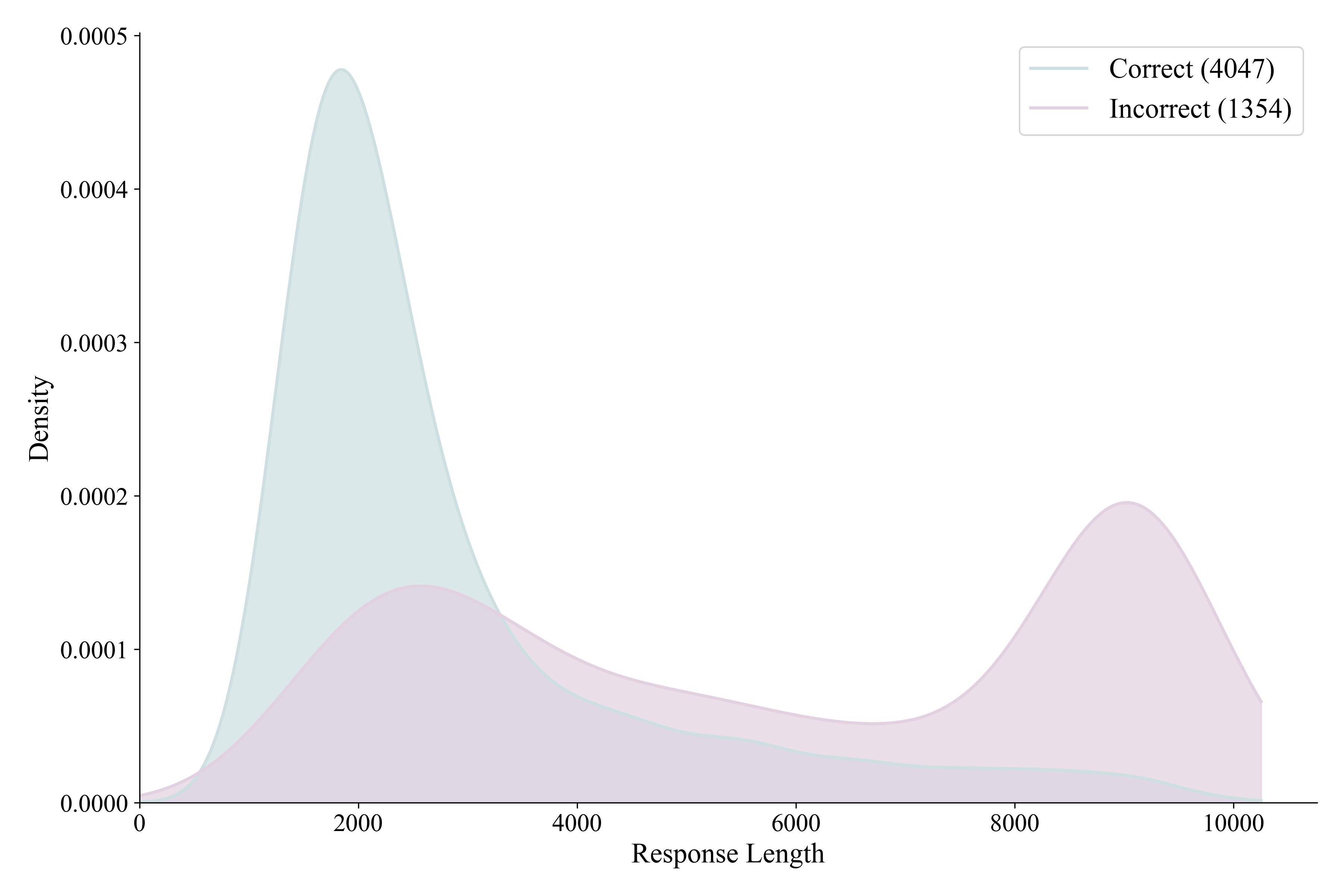}
    \subcaption{DeepSeek-R1}
  \end{subfigure}
  \begin{subfigure}[b]{0.4\textwidth}
    \centering
    \includegraphics[width=\textwidth]{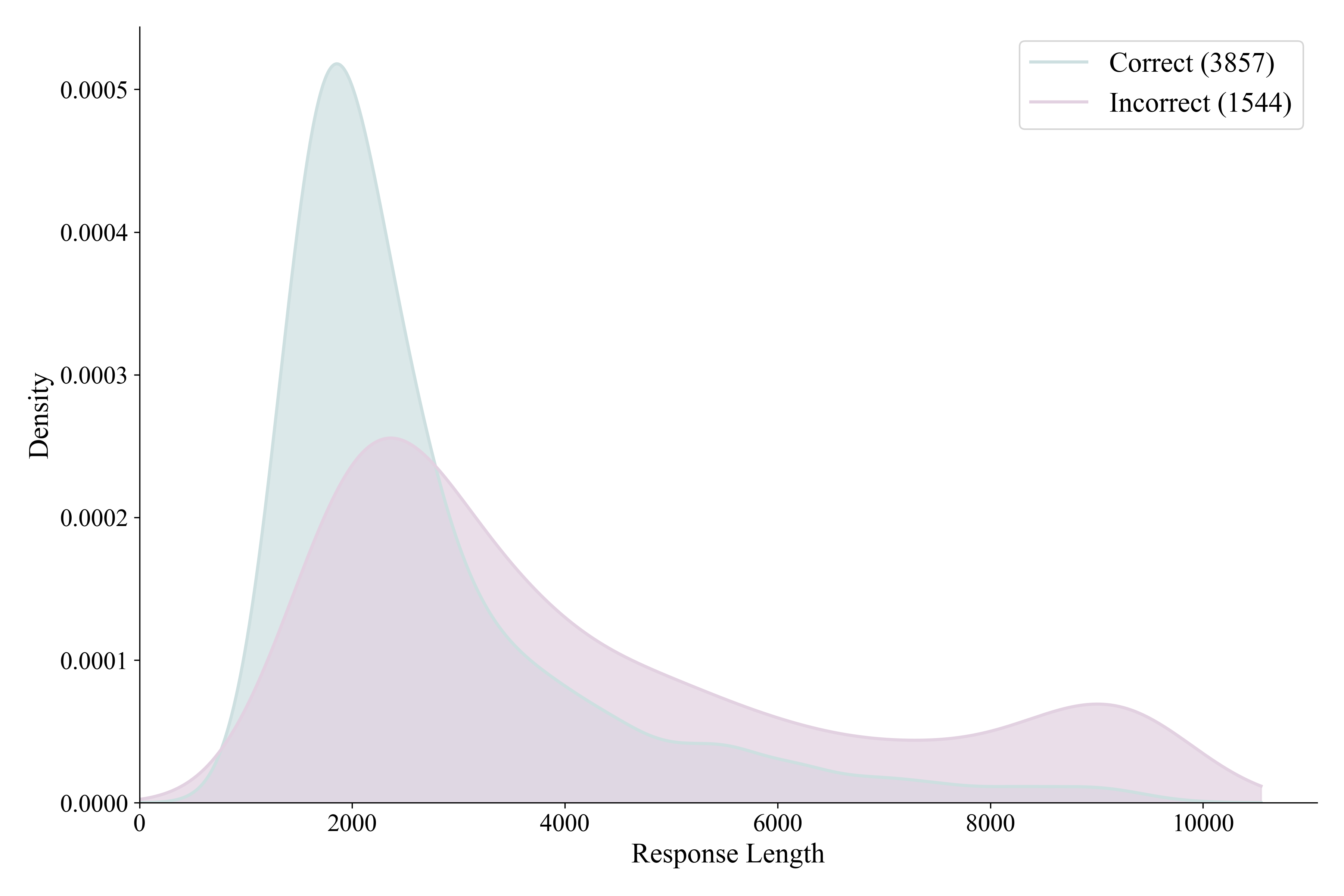}
    \subcaption{Qwen3-32B-Reasoning}
  \end{subfigure}
  \begin{subfigure}[b]{0.4\textwidth}
    \centering
    \includegraphics[width=\textwidth]{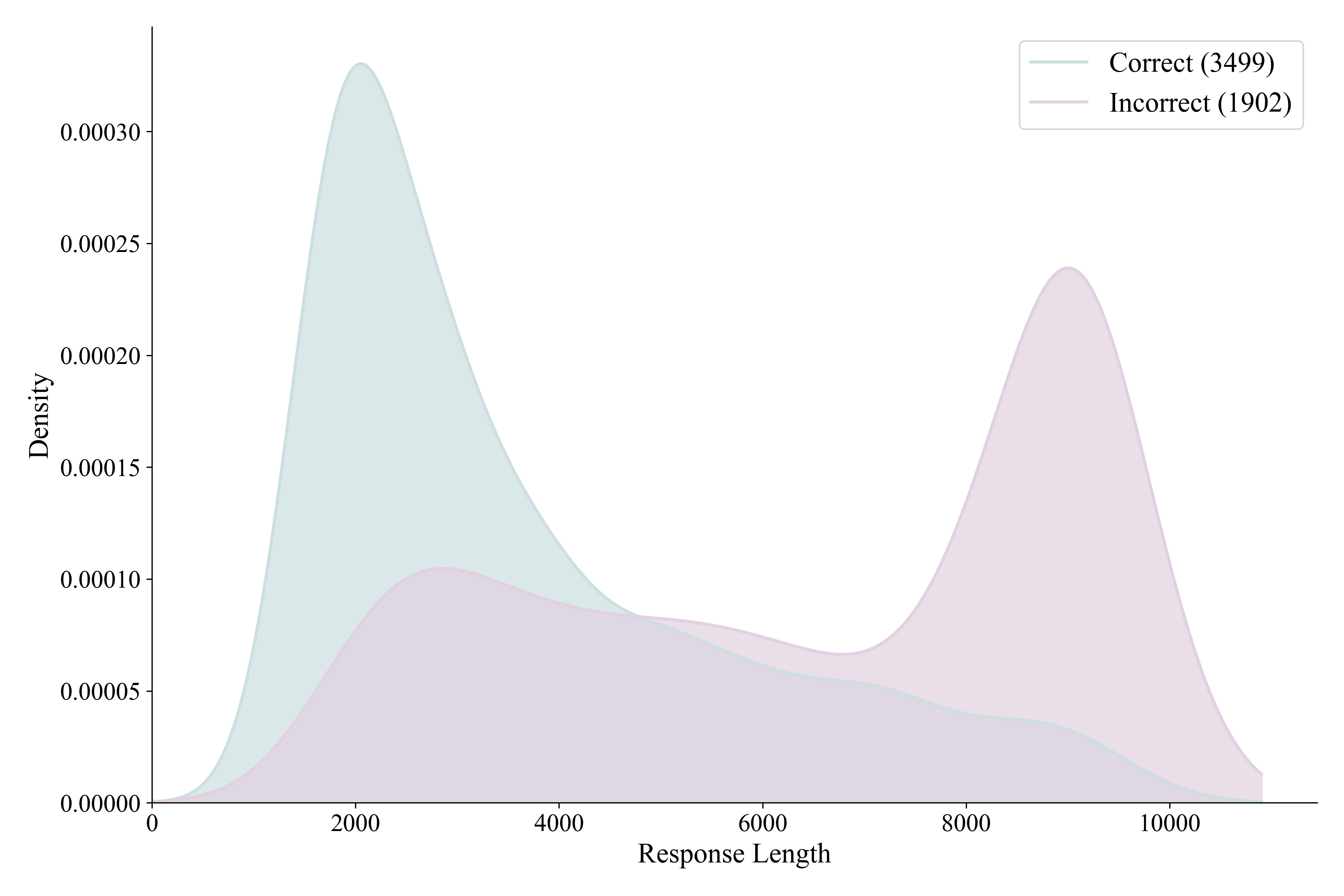}
    \subcaption{Qwen3-8B-Reasoning}
  \end{subfigure}
  \caption{Response Length Distribution on ToMATO.}
\label{app:rl_tomato}
\end{figure}

\begin{figure}[h]
  \centering
  \begin{subfigure}[b]{0.4\textwidth}
    \centering
    \includegraphics[width=\textwidth]{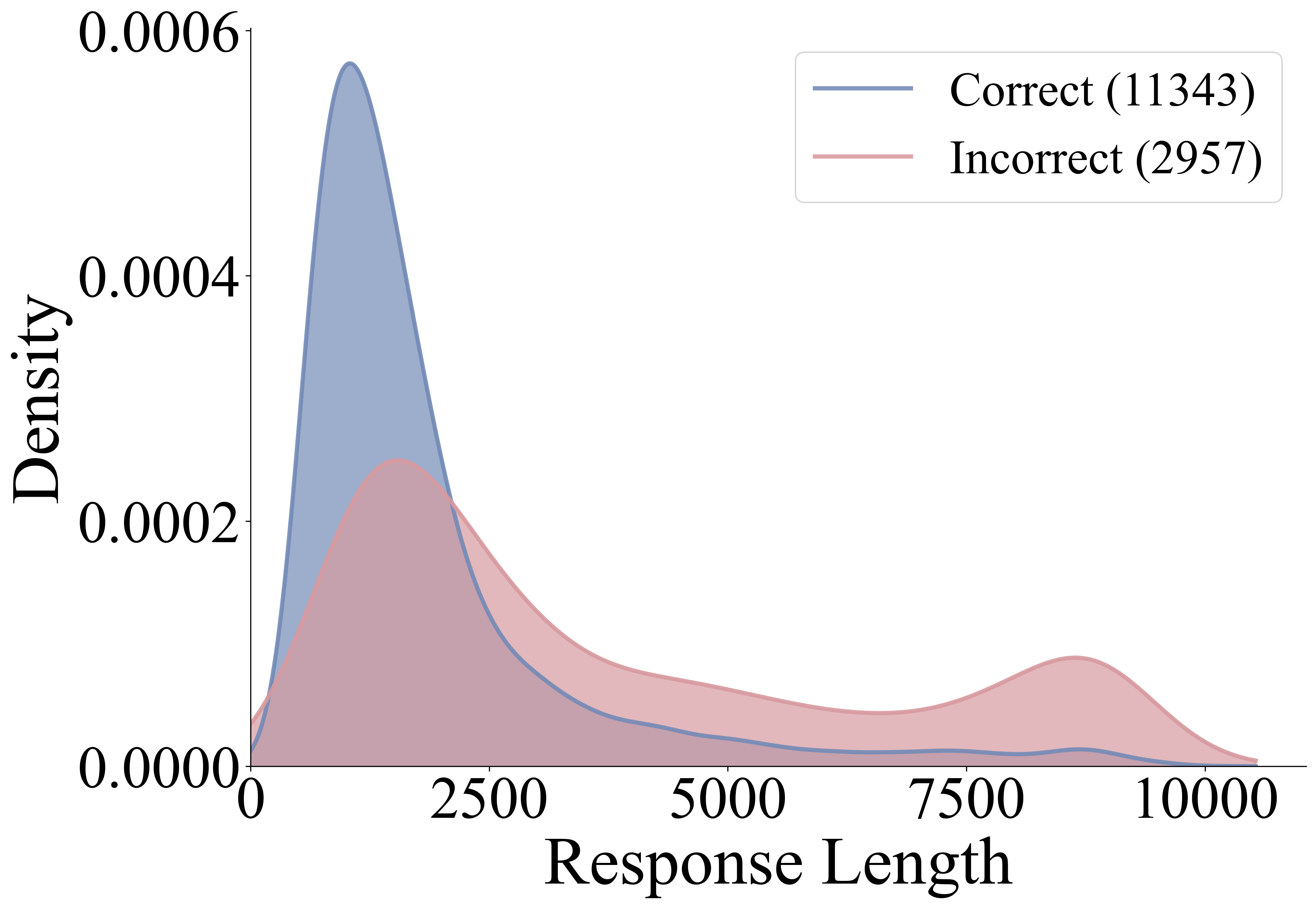}
    \subcaption{DeepSeek-R1}
  \end{subfigure}
  \begin{subfigure}[b]{0.4\textwidth}
    \centering
    \includegraphics[width=\textwidth]{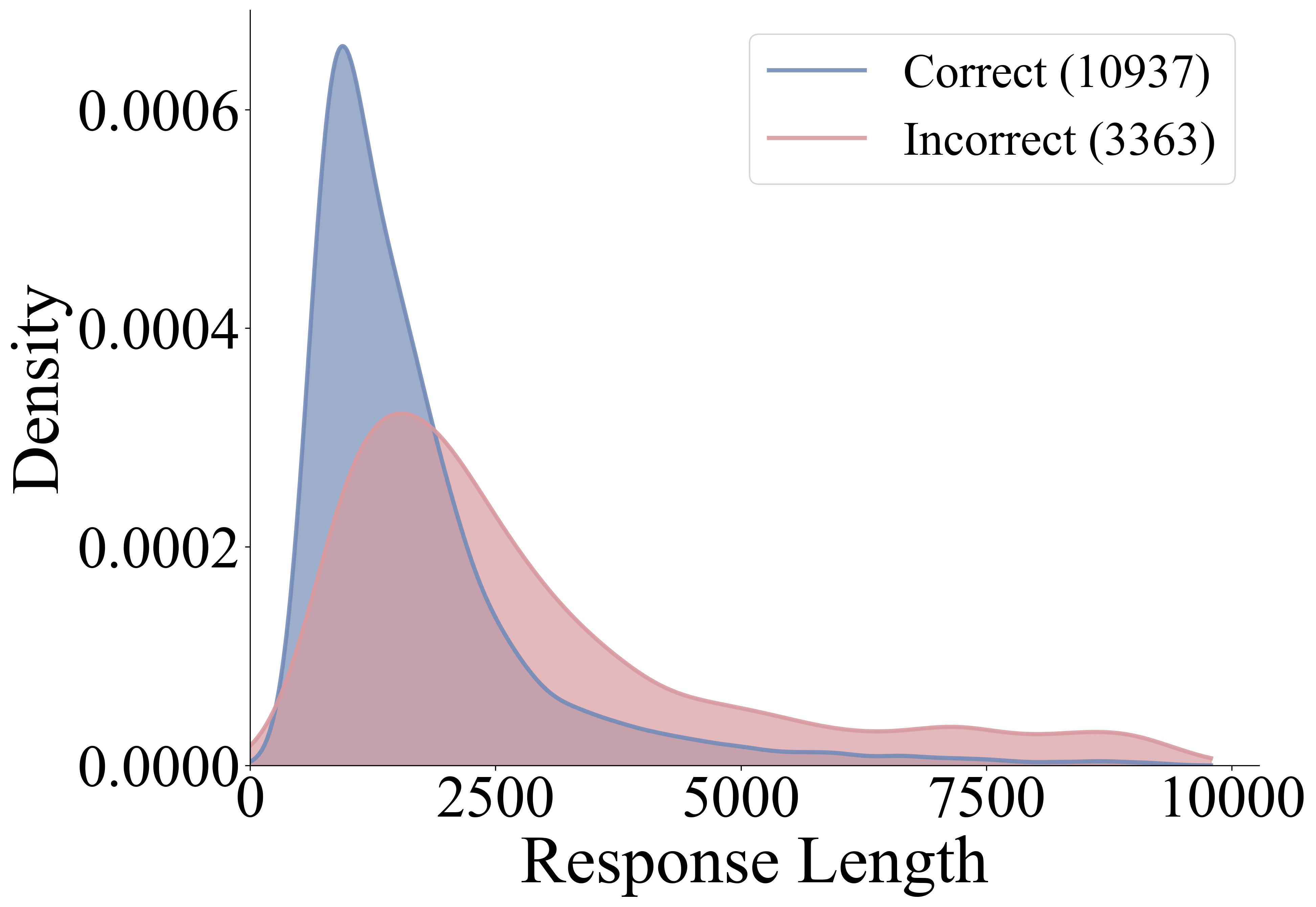}
    \subcaption{Qwen3-32B-Reasoning}
  \end{subfigure}
  \begin{subfigure}[b]{0.4\textwidth}
    \centering
    \includegraphics[width=\textwidth]{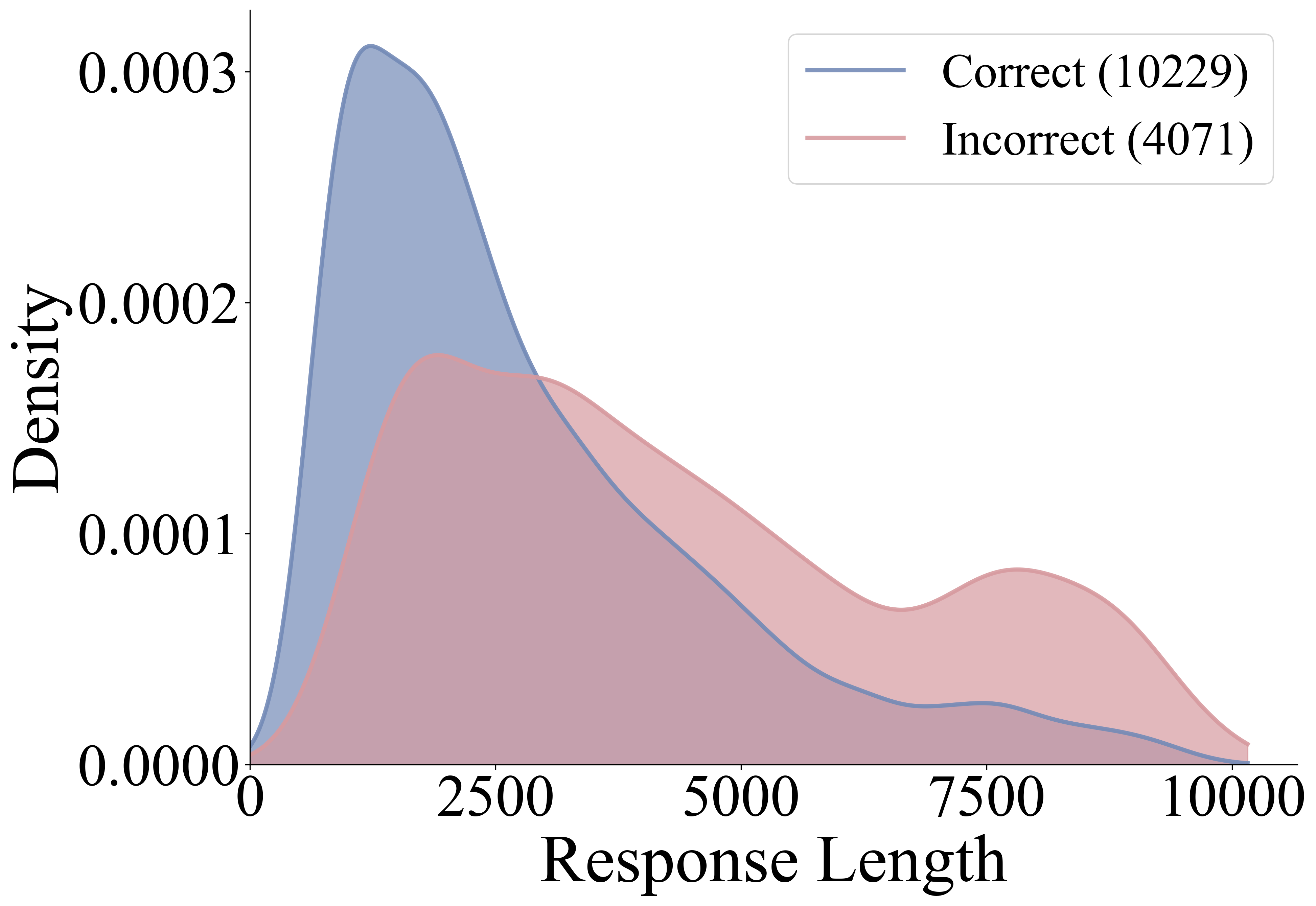}
    \subcaption{Qwen3-8B-Reasoning}
  \end{subfigure}
  \caption{Response Length Distribution on ToMBench.}
\label{app:rl_tombench}
\end{figure}

The response length distributions for the models are presented for each benchmark: HiToM in Figure \ref{app:rl_hitom}, ToMATO in Figure \ref{app:rl_tomato}, and ToMBench in Figure \ref{app:rl_tombench}.
A clear trend emerges when comparing these distributions in order of task complexity. The distinct pattern of failure we have identified—where errors cluster in an extremely high response-length region—is most pronounced on the most complex benchmark, HiToM. This effect is noticeably mitigated on ToMATO and is least apparent on ToMBench. This progression provides strong corroborating evidence for our hypothesis: the counterproductive slow thinking that leads to reasoning collapse is a failure mode that is specifically triggered and amplified by high task complexity.
We do not present GPT-o3 and GPT-o4-mini in this analysis as their genuine reasoning processes are not disclosed.

\FloatBarrier

\subsection{Order and Length}
\label{app:heat_more}
\begin{figure}[h]
  \centering
  \begin{subfigure}[b]{0.4\textwidth}
    \centering
    \includegraphics[width=\textwidth]{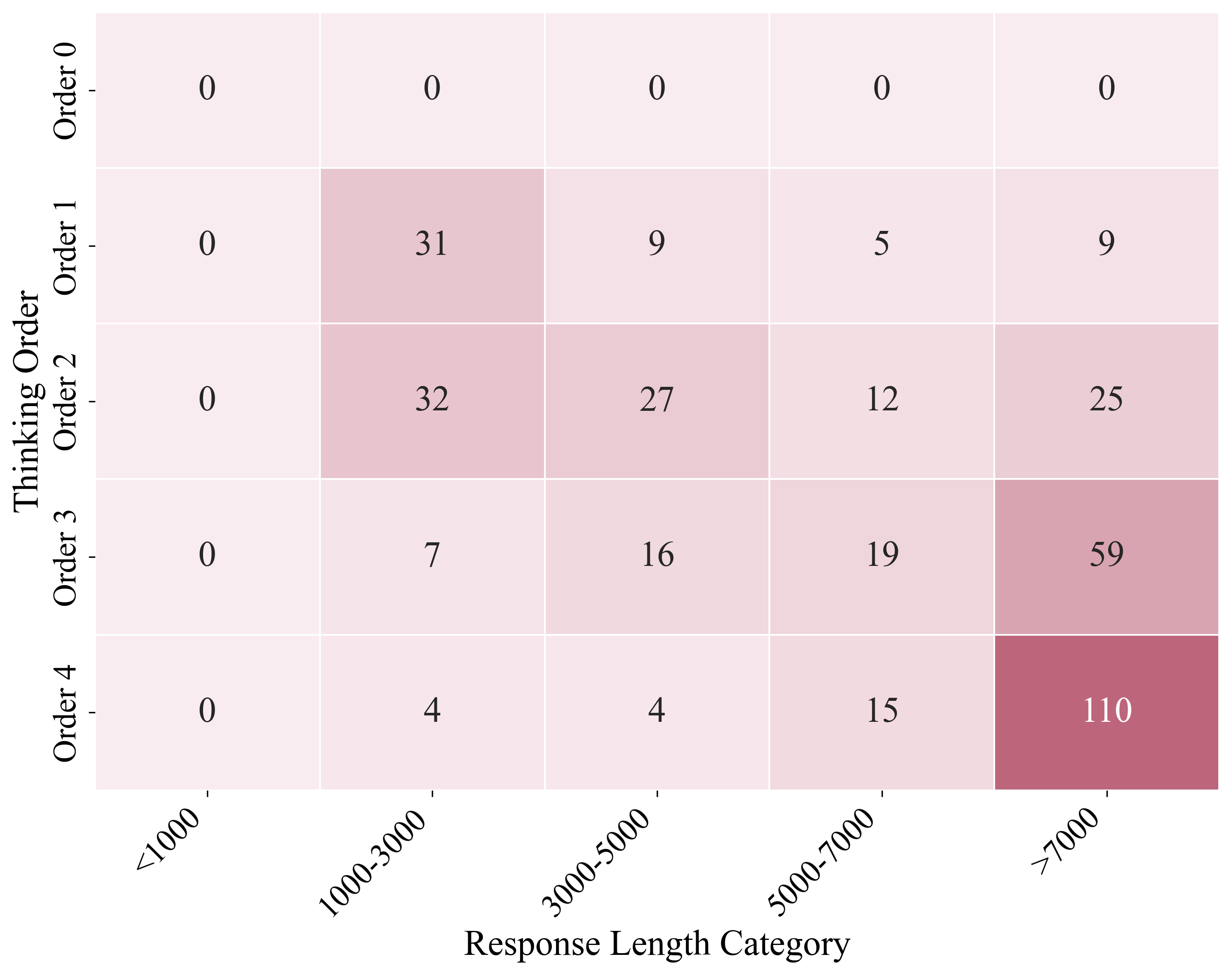}
    \subcaption{Qwen3-32B-Reasoning}
  \end{subfigure}
  \begin{subfigure}[b]{0.4\textwidth}
    \centering
    \includegraphics[width=\textwidth]{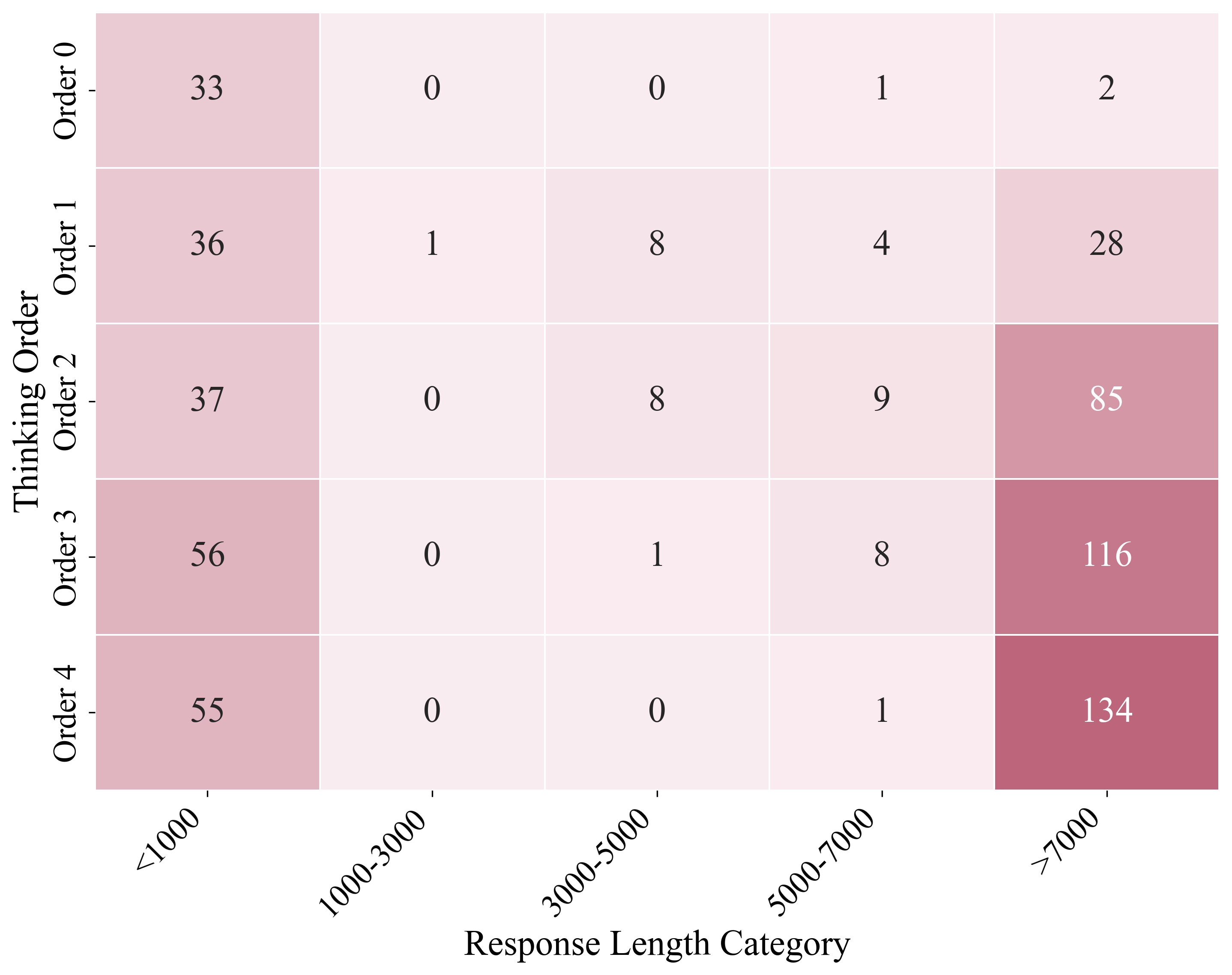}
    \subcaption{Qwen3-8B-Reasoning}
  \end{subfigure}
  \caption{Order and Length on HiToM}
\label{app:heat}
\end{figure}

We provide the heatmaps of incorrect answers on different orders and response lengths on HiToM in Figure \ref{app:heat}.
\FloatBarrier

\subsection{Reasoning Effort}
\label{app:effort_tombench}
\begin{figure}[h]
    \centering
    \includegraphics[width=0.3\linewidth]{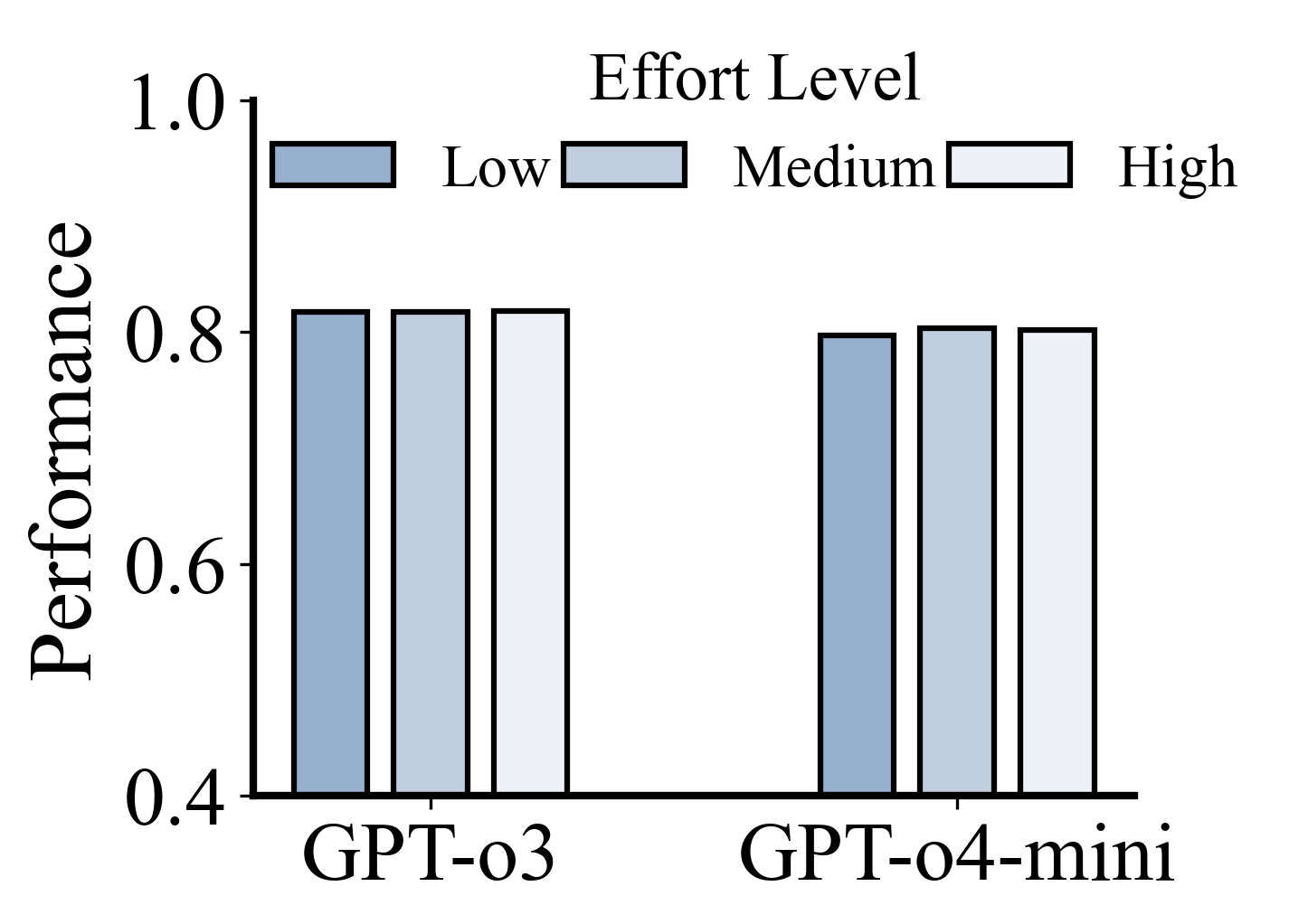}
    \caption{Resoning effort on ToMBench.}
    \label{app:rf}
\end{figure}

We provide the results of different reasoning efforts on ToMBench in Figure \ref{app:rf}. The findings are similar to those on ToMATO: increasing the reasoning effort does not lead to a significant change in performance. This reinforces our conclusion that the detrimental effects of slow thinking are specifically triggered by high task complexity. On less complex benchmarks like ToMBench and ToMATO, the reasoning collapse failure mode is not induced, and therefore, additional computational effort is neither beneficial nor harmful.
\FloatBarrier

\subsection{Token Control}
\label{app:token}
\begin{figure}
    \centering
    \begin{subfigure}[t]{0.4\textwidth}
    \centering
    \includegraphics[width=\textwidth]{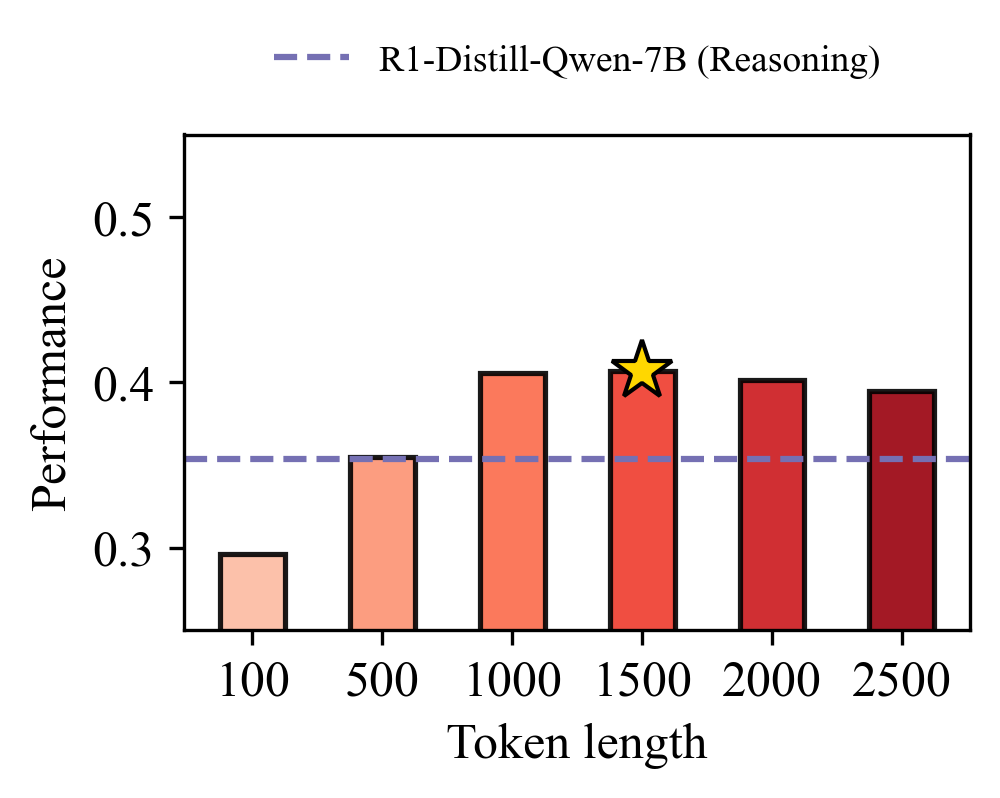}
    \caption{R1-Distill-Qwen-7B on HiToM}\label{fig:token3}
  \end{subfigure}%
  \begin{subfigure}[t]{0.4\textwidth}
    \centering
    \includegraphics[width=\textwidth]{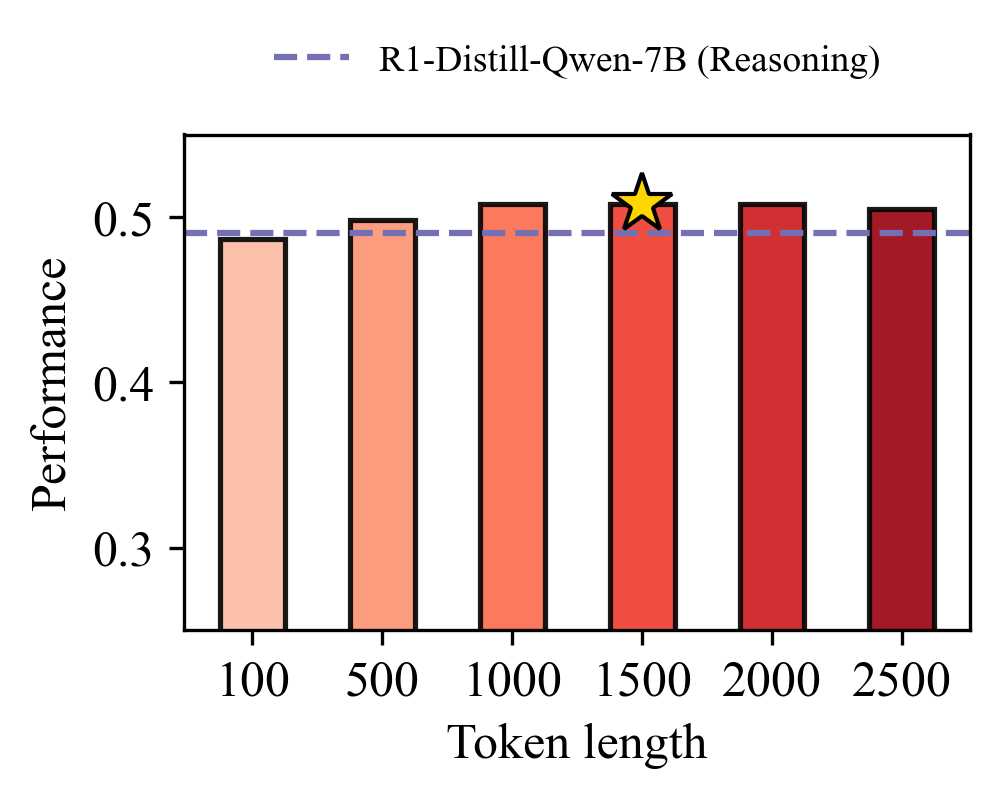}
    \caption{R1-Distill-Qwen-7B on ToMATO}\label{fig:token4}
  \end{subfigure}
    \caption{Performance comparison under different token length
limitations. The dash lines show the original model performance}
    \label{fig:r1}
\end{figure}

We provide the performance under different thinking budgets of R1-Distill-Qwen-7B in Figure \ref{fig:r1}.

\subsection{Correct Answer Overlap}
\label{app:overlap}
\begin{figure}[h]
  \centering
  \begin{subfigure}[b]{1\textwidth}
    \centering
    \includegraphics[width=\textwidth]{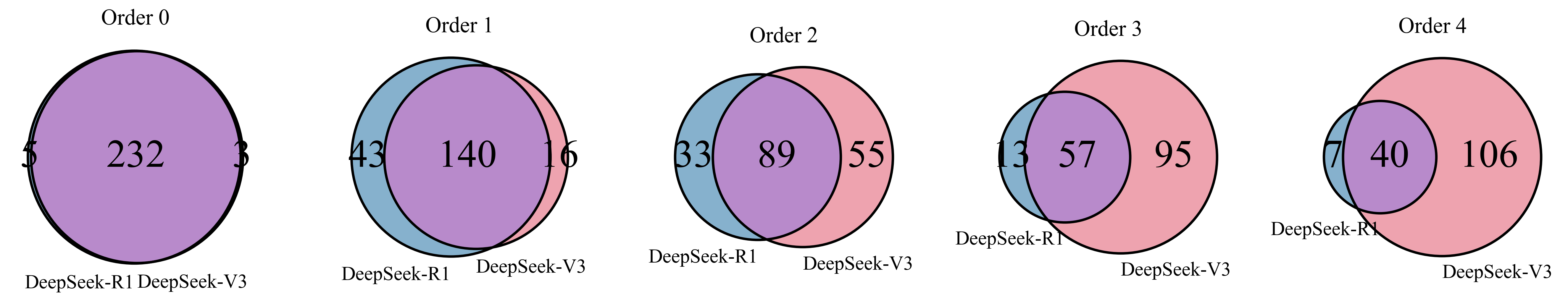}
    \subcaption{DeepSeek Family}
  \end{subfigure}
  \begin{subfigure}[b]{1\textwidth}
    \centering
    \includegraphics[width=\textwidth]{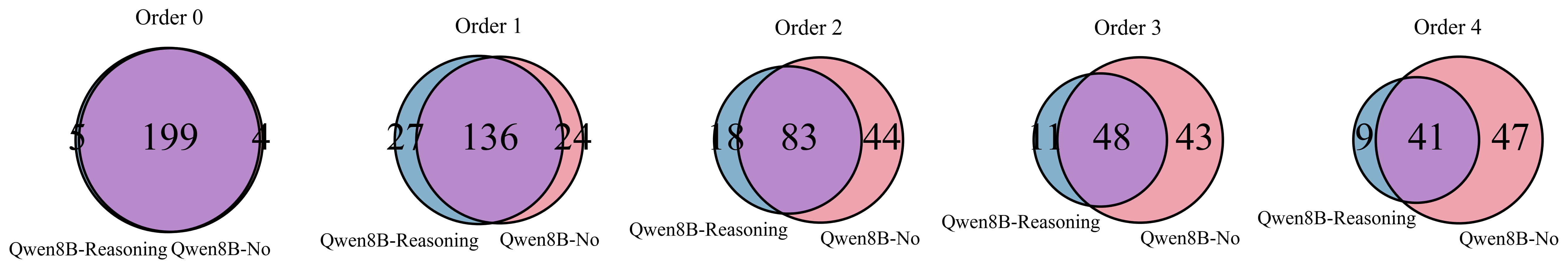}
    \subcaption{Qwen3-8B}
  \end{subfigure}
  \begin{subfigure}[b]{1\textwidth}
    \centering
    \includegraphics[width=\textwidth]{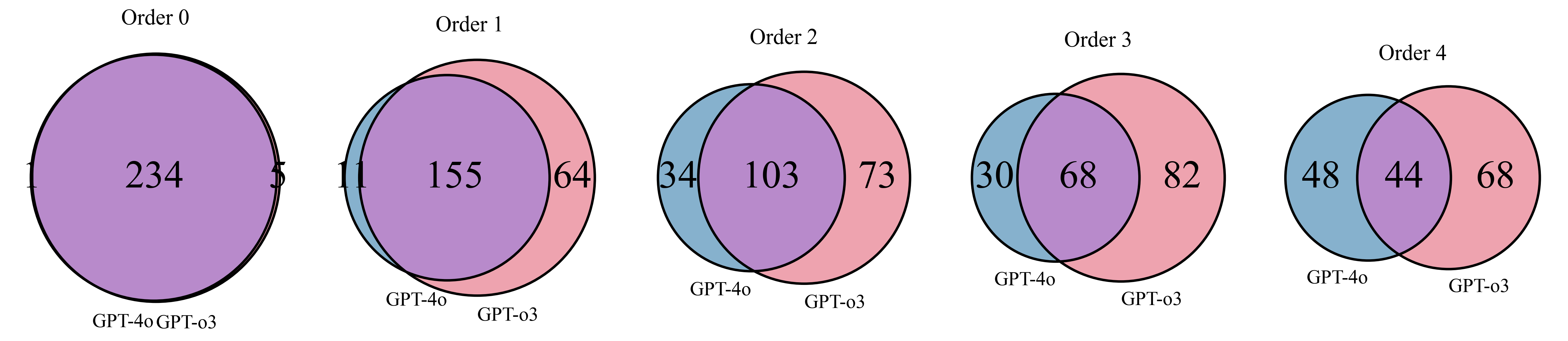}
    \subcaption{GPT Family}
  \end{subfigure}
  \caption{Correct Answer Overlap on HiToM}
\label{app:over_hi}
\end{figure}

\begin{figure}[h]
  \centering
  \begin{subfigure}[b]{0.44\textwidth}
    \centering
    \includegraphics[width=\textwidth]{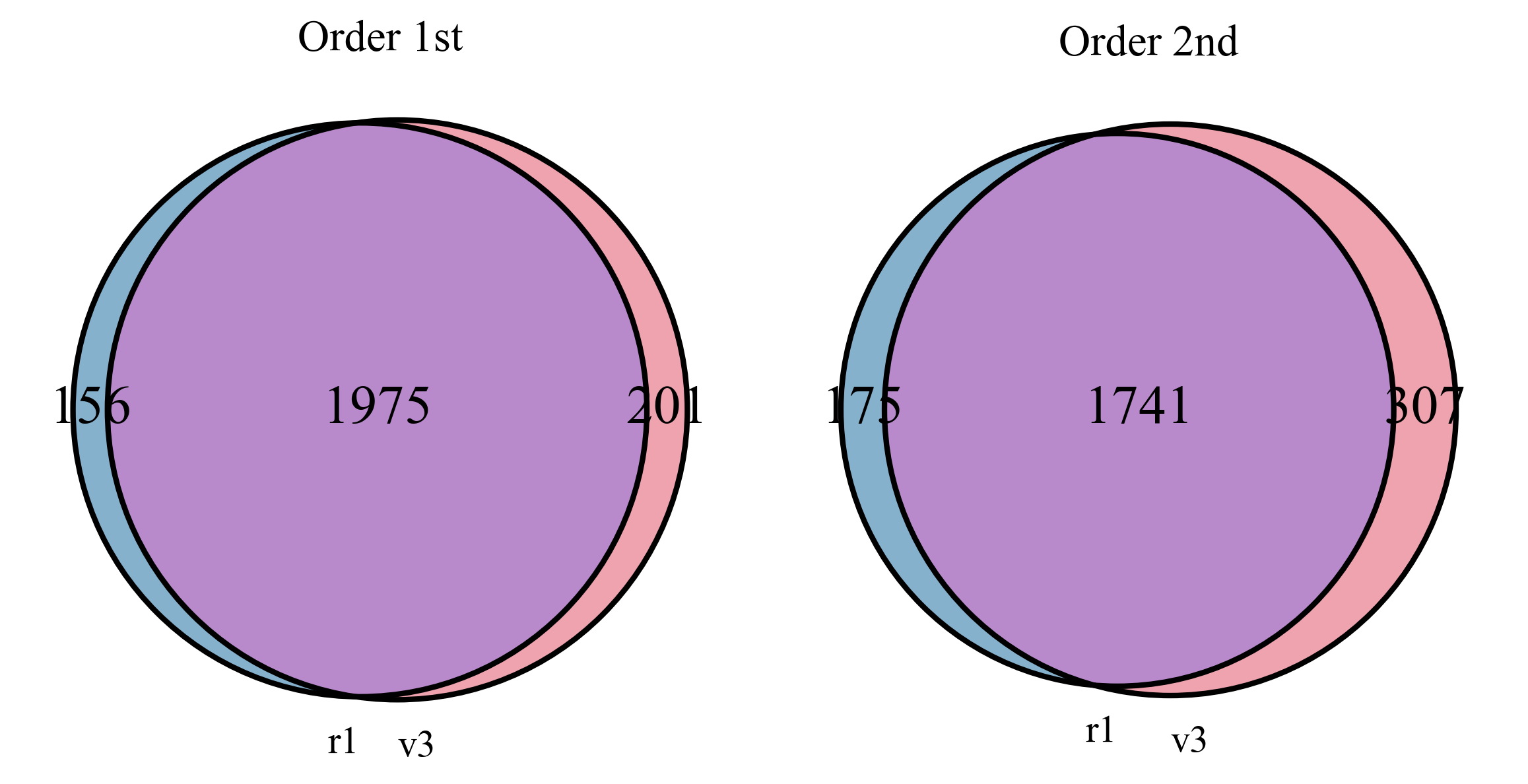}
    \subcaption{DeepSeek Family}
  \end{subfigure}
  \begin{subfigure}[b]{0.44\textwidth}
    \centering
    \includegraphics[width=\textwidth]{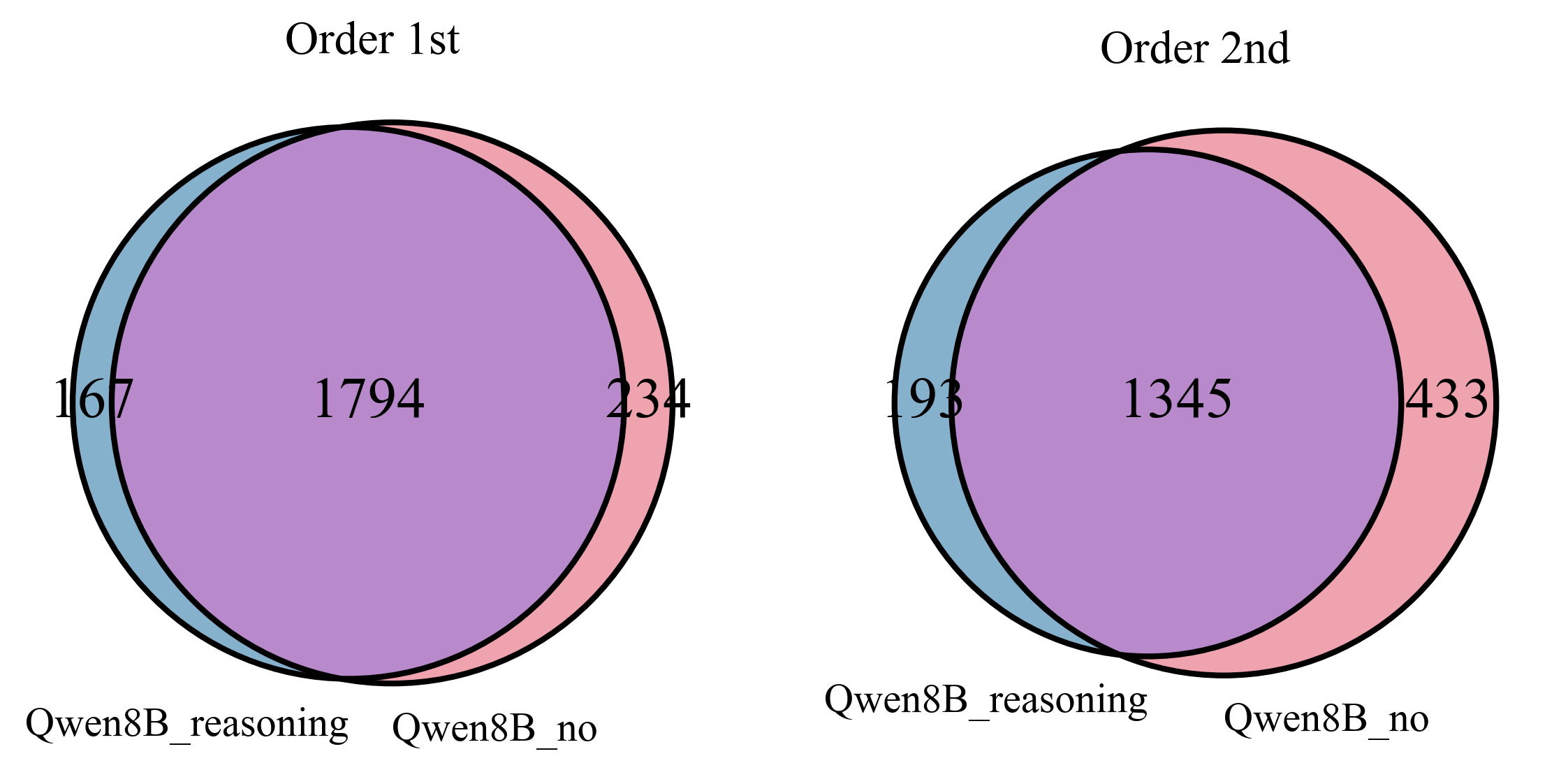}
    \subcaption{Qwen3-8B}
  \end{subfigure}
  \begin{subfigure}[b]{0.44\textwidth}
    \centering
    \includegraphics[width=\textwidth]{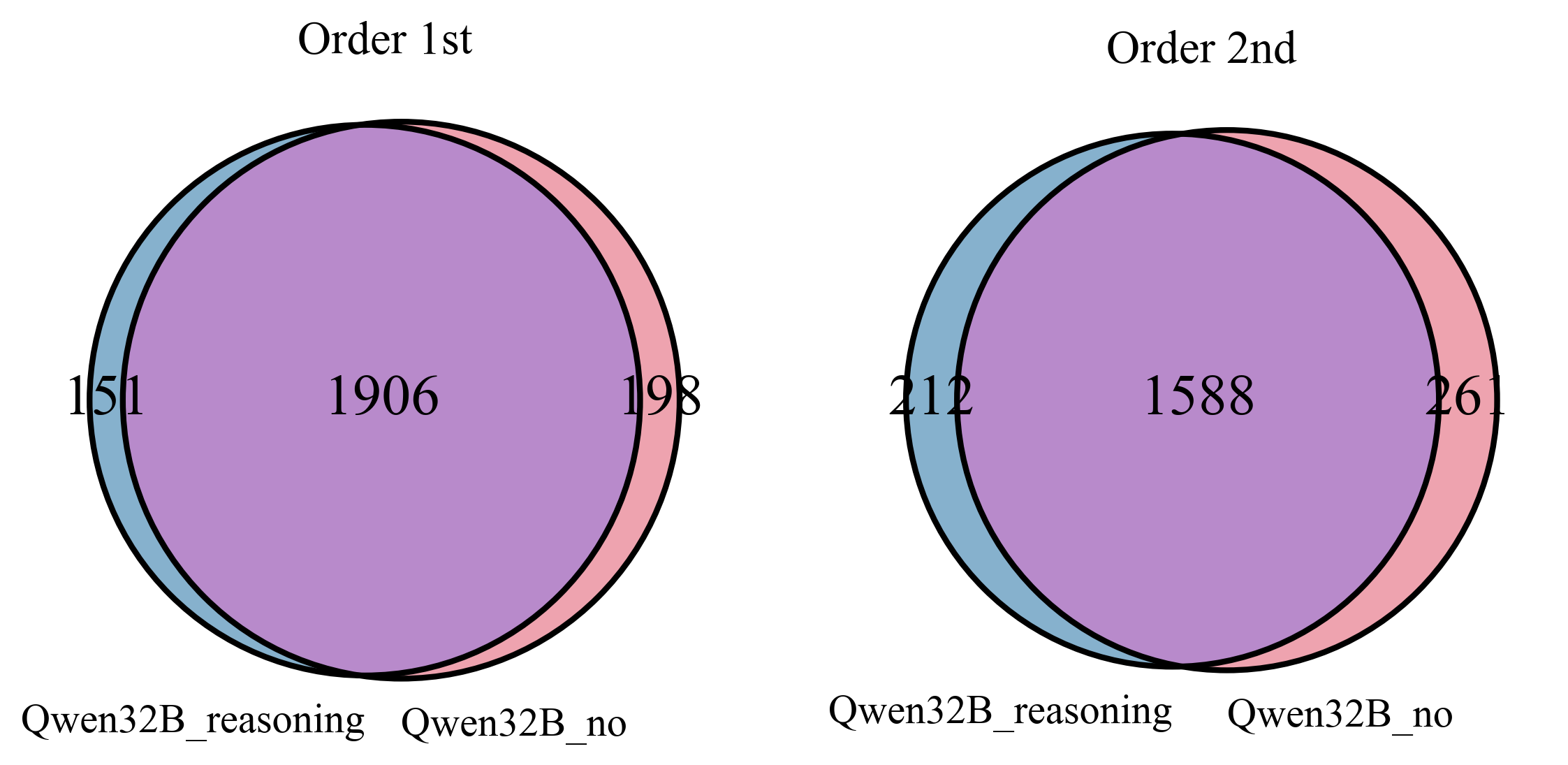}
    \subcaption{Qwen3-32B}
  \end{subfigure}
  \begin{subfigure}[b]{0.44\textwidth}
    \centering
    \includegraphics[width=\textwidth]{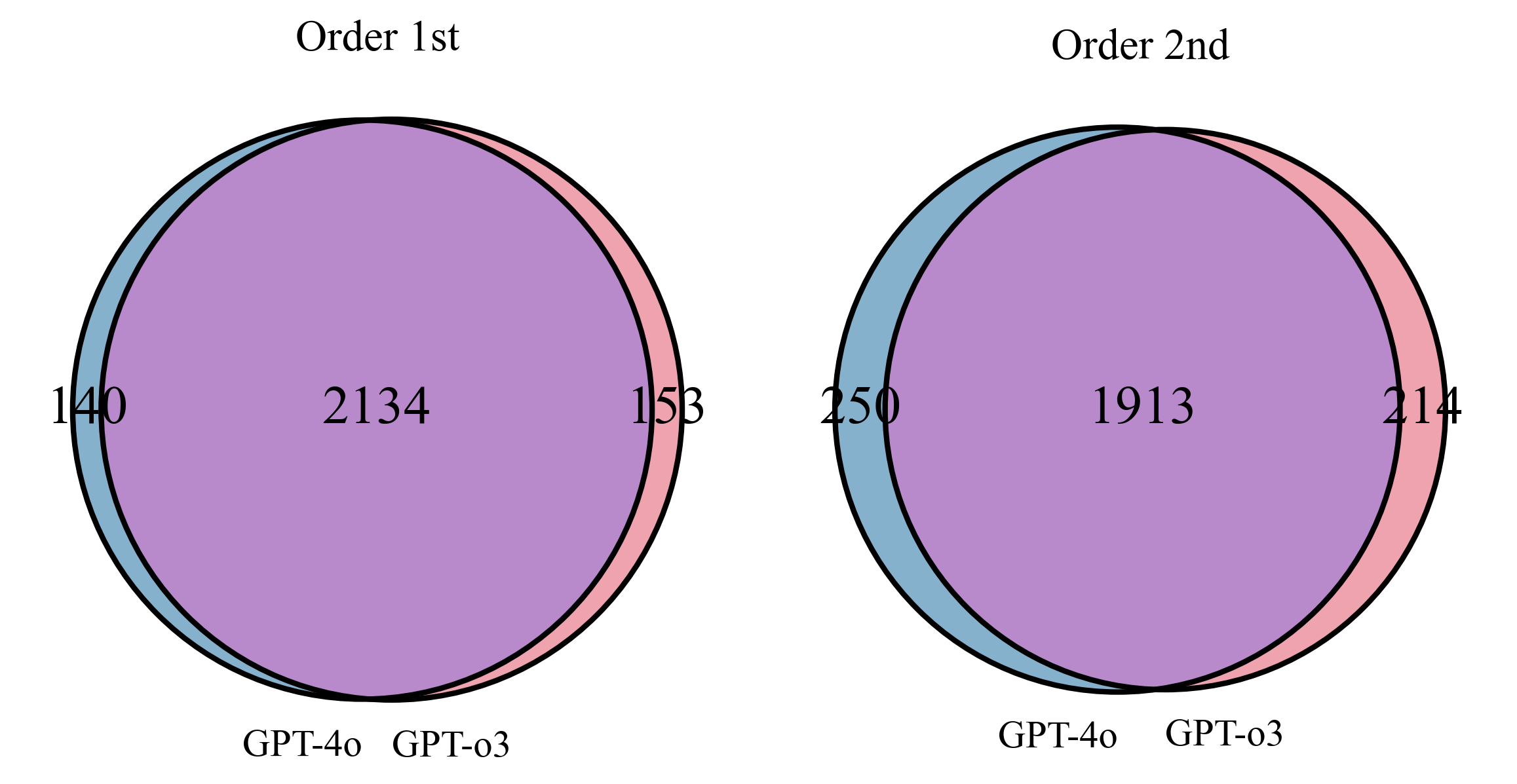}
    \subcaption{GPT Family}
  \end{subfigure}
  \caption{Correct Answer Overlap on ToMATO}
\label{app:over_to}
\end{figure}

We provide the results of correct answer overlap on HiToM in Figure \ref{app:over_hi} and on ToMATO in Figure \ref{app:over_to}.
They aligns with the observation in Figure~\ref{fig:overlap}, where the overlap between correct answers of reasoning and non-reasoning models in the same family grows when the order is higher.
The results imply the complementary advantages of two types of models.

\FloatBarrier

\section{Method}
\subsection{Pseduo-code}
\label{app:me}
We provide pseudo-code of S2F and T2M in Algorithm \ref{pc:s2f} and \ref{pc:t2m} respectively for reproducibility. Specifically, we set the threshold of ``wait'' count to 3 in our experiments. As S2F intervention requires token-level generation control, we conduct these experiments on open-source models. To analyse comprehensively, we use Qwen3 models and introduce two R1-Distill-Qwen variants.
\label{app:method}
\begin{algorithm}[h]
\caption{Slow-to-Fast Reasoning}\label{pc:s2f}
\begin{algorithmic}[1]
\STATE {\bf Input:} Prompt $x$, LLM $f_\theta$, threshold $\tau$, target token $w=\text{``wait''}$, preset sentence $S_{\text{ins}}$, max length $L_{\max}$
\STATE {\bf Output:} Generated text $Y$

\STATE $Y \gets ""$; $c \gets 0$; $finished \gets$ {\bf false}
\WHILE{not $finished$}
    \STATE $p(\cdot) \gets f_\theta(x \oplus Y)$
    \STATE $t^\star \gets \arg\max_t p(t)$
    \IF{$c \ge \tau - 1$ {\bf and} $t^\star = w$}
        \STATE $Y \gets Y \oplus S_{\text{ins}}$
        \STATE $c \gets 0$
    \ELSE
        \STATE sample $t \sim \text{Decode}(p(\cdot))$
        \STATE $Y \gets Y \oplus t$
        \IF{$t = w$}
            \STATE $c \gets c + 1$
        \ENDIF
        \IF{$t$ is EOS {\bf or} length$(Y) \ge L_{\max}$}
            \STATE $finished \gets$ {\bf true}
        \ENDIF
    \ENDIF
\ENDWHILE
\STATE {\bf return} $Y$
\end{algorithmic}
\end{algorithm}

\vspace{-0.7cm}

\begin{algorithm}[h]
\caption{Think-to-Match}\label{pc:t2m}
\begin{algorithmic}[1]
\STATE {\bf Inputs:} base prompt $x_{\text{base}}$ (prompt without options), options string $O$, LLM $f_\theta$, threshold $\tau$, target token $w=\text{``wait''}$, preset sentence $S_{\text{ins}}$, max length $L_{\max}$
\STATE {\bf Output:} Generated text $Y$
\STATE $S_{\text{full}} \gets S_{\text{ins}} \oplus O$ 
\STATE $Y \gets ""$; $c \gets 0$; $finished \gets$ {\bf false}; $inserted \gets$ {\bf false}
\WHILE{{\bf not} $finished$}
    \STATE $p(\cdot) \gets f_\theta(x_{\text{base}} \oplus Y)$
    \STATE $t^\star \gets \arg\max_t p(t)$
   
    \IF{$(c \ge \tau - 1)$ {\bf and} $(t^\star = w)$ {\bf and not} $inserted$}
        \STATE $Y \gets Y \oplus S_{\text{full}}$
        \STATE $c \gets 0$; $inserted \gets$ {\bf true}
        \STATE {\bf continue}
    \ENDIF

    \IF{$(t^\star = </think>)$ {\bf and not} $inserted$}
        \STATE $Y \gets Y \oplus S_{\text{full}}$
        \STATE $c \gets 0$; $inserted \gets$ {\bf true}
        \STATE {\bf continue}
    \ENDIF

    \STATE sample $t \sim \text{Decode}(p(\cdot))$ 
    \STATE $Y \gets Y \oplus t$
    \IF{$t = w$} \STATE $c \gets c + 1$ \ENDIF
    \IF{$t = \langle\text{END}\rangle$ {\bf or} $\mathrm{length}(Y) \ge L_{\max}$}
        \STATE $finished \gets$ {\bf true}
    \ENDIF
\ENDWHILE
\STATE {\bf return} $Y$
\end{algorithmic}
\end{algorithm}
\FloatBarrier
\subsection{Case Study}
We provide detailed cases of reasoning process of T2M on the three benchmarks.
\begin{tcolorbox}[
    breakable,
    colback=lightgray,
    width=1\textwidth,
    boxrule=0pt, 
    arc=0pt 
]
Case on \textbf{HiToM}
\newline \textbf{Story:} ``The following story happens in chronological order. You will be given a multiple-choice question and a note at the end. First give step-by-step analysis about the question. Then output the answer.
\newline 1 Sophia, Jacob, Lily, Lucas and Alexander entered the bathroom.
\newline 2 The green\_pepper is in the blue\_suitcase.
\newline 3 Sophia moved the green\_pepper to the blue\_bottle.
\newline 4 Sophia exited the bathroom.
\newline 5 Jacob made no movements and stayed in the bathroom for 1 minute.
\newline 6 Jacob exited the bathroom.
\newline 7 Lily moved the green\_pepper to the green\_pantry.
\newline 8 Lily exited the bathroom.
\newline 9 Lucas made no movements and stayed in the bathroom for 1 minute.
\newline 10 Lucas exited the bathroom.
\newline 11 Alexander moved the green\_pepper to the blue\_suitcase.
\newline 12 Alexander exited the bathroom.
\newline 13 Sophia, Jacob, Lily, Lucas and Alexander entered the waiting\_room.''
\newline
\newline \textbf{Question:} ``Where is the green\_pepper really?''
\newline
\newline \textbf{Choices:} ``A. blue\_suitcase, B. red\_bottle, C. green\_pantry, D. green\_envelope, E. blue\_bottle, F. red\_cupboard, G. green\_cupboard, H. green\_bathtub, I. green\_basket, J. green\_crate, K. blue\_crate, L. blue\_bathtub, M. blue\_cupboard, N. red\_suitcase, O. red\_container''
\newline
\newline \textbf{Response:}
\newline \texttt{<think>}
\newline Okay, let's try to figure out where the green\_pepper is. The story is in chronological order, so I need to follow each step carefully.
\newline
\newline First, the initial state: The green\_pepper is in the blue\_suitcase. That's step 2. Then, step 3 says Sophia moves it to the blue\_bottle. So after that, the green\_pepper is in the blue\_bottle. Then Sophia exits (step 4).
\newline
\newline Next, Jacob stays for a minute (step 5) and exits (step 6). No movement here, so the green\_pepper remains in the blue\_bottle. Then Lily moves it to the green\_pantry (step 7), so now it's in the green\_pantry. She exits (step 8).
\newline
\newline Then Lucas stays for a minute (step 9) and exits (step 10). No movement, so the green\_pepper is still in the green\_pantry. Then Alexander moves it to the blue\_suitcase (step 11), so now it's back in the blue\_suitcase. He exits (step 12).
\newline
\newline Finally, everyone enters the waiting\_room (step 13). The question is asking where the green\_pepper really is. The last action was Alexander moving it to the blue\_suitcase. So the final location should be the blue\_suitcase. Let me check again to make sure I didn't miss any steps.
\newline
\newline Yes, step 11 is the last movement. So the answer should be blue\_suitcase.
\newline
\newline Now I need to choose an answer based on my intuition from: A. blue\_suitcase, B. red\_bottle, C. green\_pantry, D. green\_envelope, E. blue\_bottle, F. red\_cupboard, G. green\_cupboard, H. green\_bathtub, I. green\_basket, J. green\_crate, K. blue\_crate, L. blue\_bathtub, M. blue\_cupboard, N. red\_suitcase, O. red\_container
\newline \texttt{</think>}
\newline
\newline \textbf{Answer:} A. blue\_suitcase
\end{tcolorbox}

\begin{tcolorbox}[
    breakable,
    colback=lightgray,
    width=1\textwidth,
    boxrule=0pt, 
    arc=0pt 
]
Case on \textbf{ToMATO}
\newline \textbf{Conversation:} Liam Johnson: ``Hi!''
\newline Ethan Johnson: ``Hi, how are you?''
\newline Liam Johnson: ``Ethan, I've noticed you've been a bit distant lately, is everything okay?''
\newline Ethan Johnson: ``Yeah, everything's fine, just been really busy with the restaurant and some personal stuff, you know how it is.''
\newline Liam Johnson: ``Come on, Ethan, we've been friends for years, I can tell when something's bothering you. What's really going on?''
\newline Ethan Johnson: ``Alright, alright... to be honest, I've been having some trouble making ends meet, the restaurant's been slow and I'm falling behind on some bills, but I'll figure it out, I always do.''
\newline Liam Johnson: ``Ethan, you're not going to 'figure it out' on your own, you need a plan and some guidance. Let me take a look at your finances and we can go over some options together, okay?''
\newline Ethan Johnson: ``Man, I appreciate that, Liam, but I don't want to be a burden, I can handle it, I just need to get my head out of my... okay, fine, yeah, maybe a fresh set of eyes would help, thanks.''
\newline Liam Johnson: ``Alright, let's get down to business then. I'll take a look at your books and we'll get a plan in place. But next time, don't be so stubborn and come to me sooner, got it?''
\newline Ethan Johnson: ``Ha! Yeah, yeah, got it, I'll try to work on that, thanks, Liam. And thanks again, man, I owe you one... or a few thousand, it seems.''
\newline Liam Johnson: ``Yeah, yeah, just pay me back in free meals at the restaurant, and we'll call it square. But seriously, let's take a close look at your expenses and see where we can cut back, okay?''
\newline Ethan Johnson: ``Aww, man, you're a lifesaver, Liam. Yeah, let's dig in and see where I can tighten the belt, I'm sure there's some fat to trim. And don't worry, you'll be eating like a king at my place for a long time to come.''
\newline Liam Johnson: ``Alright, let's get to work then. I'll start going over the numbers and we can discuss some options. And don't make promises you can't keep, just focus on getting back on your feet, okay?''
\newline Ethan Johnson: ``Yeah, yeah, got it, thanks for keeping me grounded, Liam. I'll let you do your magic with the numbers, and I'll focus on getting the restaurant back on track. You're right, one step at a time.''
\newline Liam Johnson: ``That's the spirit, Ethan. Just remember, I'm not a miracle worker, you need to be effort too. Let's schedule a follow-up meeting for next week and we'll reassess where we're at, sound good?''
\newline Ethan Johnson: ``Absolutely, sounds like a plan, Liam. I'll make sure to have all my ducks in a row by then. And thanks again, man, just knowing you've got my back makes me feel a hell of a lot better.''
\newline
\newline \textbf{a0:} ``He feels a sense of dry amusement at Ethan's joke, but also a bit of unease about the seriousness of his financial situation''
\newline \textbf{a1:} ``He feels a bit frustrated that Ethan is not being open about his struggles, and he's worried that he might be hiding something''
\newline \textbf{a2:} ``He feels a bit anxious about bringing up the sensitive topic of Ethan's financial struggles, but he want to help his friend''
\newline \textbf{a3:} ``He feels a sense of relief that Ethan is finally opening up and accepting help, but also a bit of irritation that he didn't listen to him sooner''
\newline
\newline \textbf{q:} ``How does Liam feel when he says ``Come on, Ethan, we've been friends for years, I can tell when something's bothering you. What's really going on?"
\newline
\newline \textbf{response:}
\newline \texttt{<think>}
\newline Okay, let's see. The question is about how Liam feels when he says that specific line. Let me look at the conversation again.
\newline
\newline So, Liam starts by greeting Ethan. Then Ethan says he's been busy with the restaurant and personal stuff. Liam responds with ``Come on, Ethan... What's really going on?'' That line seems like Liam is pushing Ethan to be honest.
\newline
\newline Earlier, Ethan had said he's been busy, but Liam is pointing out that they've been friends for years and he can tell something's up. So Liam is probably feeling concerned or worried about Ethan. He's trying to get Ethan to open up because he's noticed something's wrong.
\newline
\newline Looking at the options, even though the options aren't listed here, the possible answers would likely be emotions like concern, frustration, patience, or maybe anger. But based on the context, Liam is a friend trying to help, so it's more about concern and wanting to support Ethan. He's not angry, but rather trying to get Ethan to talk. So the answer is probably that Liam feels concerned or worried.
\newline
\newline Now I need to choose an answer based on my intuition from: [A] He feels a sense of dry amusement at Ethan's joke, but also a bit of unease about the seriousness of his financial situation
\newline [B] He feels a bit frustrated that Ethan is not being open about his struggles, and he's worried that he might be hiding something
\newline [C] He feels a bit anxious about bringing up the sensitive topic of Ethan's financial struggles, but he want to help his friend
\newline [D] He feels a sense of relief that Ethan is finally opening up and accepting help, but also a bit of irritation that he didn't listen to him sooner
\newline \texttt{</think>}
\newline
\newline \textbf{Answer:} B
\end{tcolorbox}

\begin{tcolorbox}[
    breakable,
    colback=lightgray,
    width=1\textwidth,
    boxrule=0pt, 
    arc=0pt 
]
Case on \textbf{ToMBench}
\newline\textbf{STORY:} ``Zhang Wei is the performance champion of the department, but he encounters some challenges in his recent project, which causes his performance to be not good. Li Yun is the department manager, he always supports Zhang Wei, hoping to fight for the best evaluation for him in the year-end review. Chen Tao is Zhang Wei's colleague, he also hopes to get promoted in the year-end review. This afternoon, Li Yun quickly browses Zhang Wei's review form in the office. When Zhang Wei passes by Li Yun's office, Li Yun looks up at him and nods to him. Zhang Wei gives a slight smile to Li Yun and continues to walk past. Chen Tao sees all this in the tea room not far away.''
\newline
\newline \textbf{QUESTION:} ``Why does Li Yun nod to Zhang Wei?''
\newline
\newline \textbf{OPTION-A:} ``Li Yun just sees Zhang Wei's review form and thinks his performance is good.''
\newline \textbf{OPTION-B:} ``Li Yun wants to tell Zhang Wei that he gets a high evaluation.''
\newline \textbf{OPTION-C:} ``Li Yun is greeting Zhang Wei.''
\newline \textbf{OPTION-D:} ``Li Yun nods to Zhang Wei unintentionally.''
\newline
\newline \textbf{ANSWER:} ``B''
\newline
\newline \textbf{map:} \{``A'': ``D'', ``B'': ``C'', ``C'': ``A'', ``D'': ``B''\}
\newline
\newline \textbf{output:}
\newline \texttt{<think>}
\newline Okay, let's try to figure out why Li Yun nodded to Zhang Wei. So, the story says that Zhang Wei is the performance champion of the department, but he's having some challenges in his recent project, which is affecting his performance. Li Yun is the department manager who supports Zhang Wei and wants to fight for the best evaluation for him in the year-end review. Chen Tao is another colleague who also wants to get promoted.
\newline
\newline Now, the key moment is when Li Yun quickly browses Zhang Wei's review form in the office. Then, when Zhang Wei passes by Li Yun's office, Li Yun nods at him, and Zhang Wei gives a slight smile and continues walking. Chen Tao sees this from the tea room.
\newline
\newline So, why would Li Yun nod to Zhang Wei? Let's think. Since Li Yun is the manager and wants to help Zhang Wei get the best evaluation, maybe he's trying to show support. But why nod specifically? Maybe because he just looked at Zhang Wei's review form. If he's looking at the form, perhaps he's checking Zhang Wei's performance or the evaluation. But since Zhang Wei is having challenges, maybe Li Yun is trying to encourage him. Alternatively, maybe Li Yun is trying to signal that he's aware of Zhang Wei's situation and is supporting him.
\newline
\newline Alternatively, could it be that Li Yun is trying to get Zhang Wei to know that he's being considered for a good evaluation? Or maybe Li Yun is trying to make sure that Zhang Wei's performance isn't overlooked despite the recent challenges. The nod could be a gesture of support or a reminder that Li Yun is on his side.
\newline
\newline Another angle: since Chen Tao is also hoping for promotion, maybe there's some competition. But the question is about Li Yun's action towards Zhang Wei. The story says Li Yun always supports Zhang Wei and wants to fight for the best evaluation. So the nod is likely a supportive gesture.
\newline
\newline But why would Li Yun look at the review form before nodding? Maybe he's checking the form and wants to confirm that Zhang Wei's performance is being considered, or maybe he's trying to influence the evaluation. The nod could be a way to communicate that he's aware of Zhang Wei's situation and is advocating for him.
\newline
\newline Alternatively, maybe Li Yun is trying to get Zhang Wei to acknowledge his support, but the story says Zhang Wei just gives a slight smile and continues. So the nod is probably a sign of support or encouragement.
\newline
\newline Looking at the answer options, even though they aren't listed here, the options would likely be something like: A. To show support, B. To check his performance, C. To compete with Chen Tao, D. To inform about the review. But since the options aren't given, based on the story, the most logical answer is that Li Yun is supporting Zhang Wei, so the answer would be something like A. But since the user hasn't provided the options, I have to assume that the answer is based on the given context. The answer is likely related to support or advocacy for Zhang Wei's evaluation. So the answer index would be the one that matches that reasoning.
\newline
\newline Now I need to choose an answer based on my intuition from: A. Li Yun nods to Zhang Wei unintentionally. B. Li Yun is greeting Zhang Wei. C. Li Yun just sees Zhang Wei's review form and thinks his performance is good. D. Li Yun wants to tell Zhang Wei that he gets a high evaluation.
\newline \texttt{</think>}
\newline
\newline \textbf{Final Answer:} D
\end{tcolorbox}

\end{document}